\documentclass[a4paper,fleqn]{cas-dc}
 
\usepackage[authoryear]{natbib}
\usepackage{adjustbox}
\usepackage{chngcntr}
\usepackage{placeins}
\usepackage{float}

\def\tsc#1{\csdef{#1}{\textsc{\lowercase{#1}}\xspace}}
\tsc{WGM}
\tsc{QE}

\begin{document}
\let\WriteBookmarks\relax
\def\floatpagepagefraction{1}
\def\textpagefraction{.001}

\shorttitle{Cutting-edge 3D reconstruction solutions for underwater coral reef images: A review and comparison}    

\shortauthors{Zhong et al.}  

\title [mode = title]{Cutting-edge 3D reconstruction solutions for underwater coral reef images: A review and comparison}  



%

\author[1]{Jiageng Zhong}
\affiliation[1]{institute={State Key Laboratory of Information Engineering in Surveying Mapping and Remote Sensing,},
	organization={Wuhan University},
            city={Wuhan},
            state={Hubei},
            country={China}}

\author[1,2,3]{Ming Li}
\ead{lisouming@whu.edu.cn}
\cormark[1]

\affiliation[2]{institute={Institute of Geodesy and Photogrammetry,},
	organization={ETH Zurich},
            city={Zurich},
            country={Switzerland}}

\affiliation[3]{institute={School of Robotics,},
	organization={Wuhan University},
	city={Wuhan},
	state={Hubei},
	country={China}}
	
\author[2]{Armin Gruen}

\author[2]{Konrad Schindler}

\author[4]{Xuan Liao}

\author[5]{Qinghua Guo}

\affiliation[4]{institute={Department of Land Surveying and Geo-Informatics,},
	organization={The Hong Kong Polytechnic University},
	city={Hong Kong},
	country={China}}

\affiliation[5]{institute={Institute of Remote Sensing and Geographic Information System, School of Earth and Space Sciences,},
	organization={Peking University},
	city={Beijing},
	country={China}}

\cortext[1]{Corresponding author}



\begin{abstract}
	Corals serve as the foundational habitat-building organisms within reef ecosystems, constructing extensive structures that extend over vast distances. However, their inherent fragility and vulnerability to various threats render them susceptible to significant damage and destruction. The application of advanced 3D reconstruction technologies for high-quality modeling is crucial for preserving them. These technologies help scientists to accurately document and monitor the state of coral reefs, including their structure, species distribution and changes over time. Photogrammetry-based approaches stand out among existing solutions, especially with recent advancements in underwater videography, photogrammetric computer vision, and machine learning. Despite continuous progress in image-based 3D reconstruction techniques, there remains a lack of systematic reviews and comprehensive evaluations of cutting-edge solutions specifically applied to underwater coral reef images. The emerging advanced methods may have difficulty coping with underwater imaging environments, complex coral structures, and computational resource constraints. They need to be reviewed and evaluated to bridge the gap between many cutting-edge technical studies and practical applications. This paper focuses on the two critical stages of these approaches: camera pose estimation and dense surface reconstruction. We systematically review and summarize classical and emerging methods, conducting comprehensive evaluations through real-world and simulated datasets. Based on our findings, we offer reference recommendations and discuss the development potential and challenges of existing approaches in depth. This work equips scientists and managers with a technical foundation and practical guidance for processing underwater coral reef images for 3D reconstruction. These tools facilitate the acquisition of accurate data, enhancing our understanding of the complex coral reef ecosystems while minimizing disturbances to these sensitive habitats, ultimately supporting coral reef conservation and restoration efforts.
\end{abstract}




\begin{keywords}
 Coral reefs \sep Underwater photogrammetry \sep 3D reconstruction \sep Computer vision \sep Deep learning
\end{keywords}

\maketitle

\section{Introduction}\label{}

Coral reefs are distinguished as highly complex ecosystems in warm tropical and subtropical oceans, renowned for their exceptional biodiversity, intricate structural formations, and remarkably high primary productivity (\cite{mellin2022safeguarding105}). Despite covering less than 0.1\% of the ocean’s surface, tropical reefs support approximately one-quarter to one-third of all marine species  (\cite{jones2022climatic109,plaisance2011diversity110}). However, these ecosystems are among the most vulnerable to global climate change, primarily due to the thermal sensitivity of reef-building corals, which are prone to bleaching and even death as ocean temperatures rise  (\cite{hoegh2007coral106}). Furthermore, coral reefs face significant threats from local stressors, including water pollution, intensified fishing practices, resource extraction, and coastal development  (\cite{hughes2017global107,carlson2019land108,morrison2020advancing111}). Between 2009 and 2018, approximately 14\% of coral reefs were lost globally, with projections indicating that under a high global warming emissions scenario, up to 99\% of coral reefs could experience severe bleaching events within the twenty-first century  (\cite{robinson2023global112}). These trends underscore that coral reefs and their associated fish communities face severe survival challenges. Given the critical ecological and economic importance of coral reefs, it is more urgent than ever to address and reverse the threats confronting these vulnerable ecosystems. The metabolic processes of coral colonies, including photosynthesis, respiration, calcium carbonate deposition, and reproduction, are significantly influenced by their physical characteristics, such as overall shape and topographic complexity  (\cite{pac1978coral113,burns2015integrating114}). Therefore, accurate evaluation of these physical attributes, especially through three-dimensional (3D) metrics, is essential for deepening our understanding of coral biology, as well as assessing habitat availability, biogenic flux, and overall reef productivity. It is necessary to employ advanced survey techniques for mapping, monitoring, and modeling coral reef habitats.

Coral reef surveys are conducted using a variety of platforms, including satellites, aerial systems, vessels, underwater vehicles, and manual in-situ methods (\cite{collin2018very7,price2019using16,rossi2020detecting18,costa2009comparative132}). Manual surveys are labor-intensive and limited by their spatial and temporal scope. Satellite and aerial platforms equipped with cameras or LiDAR systems enable large-scale and frequent monitoring (\cite{casella2017mapping4,collin2018very7}). However, due to their limited spatial resolution and susceptibility to surface water interference, they are generally insufficient for capturing the fine-scale structural complexity of coral reefs. Vessel-based approaches typically employ sonar, which is effective for wide-area mapping but less suited for shallow waters and incapable of capturing spectral or color information (\cite{costa2009comparative132}). Underwater LiDAR, which is capable of close-range, high-precision surveying of coral formations (\cite{vogler2019high133}), faces challenges including high costs, operational complexity, and limited capacity for capturing color information. The emergence of vision-based underwater observation techniques has allowed for the collection of high-resolution images from close distances at low cost. This advancement facilitates the use of emerging image-based 3D reconstruction methods to produce high-accuracy and high-resolution 3D models of seabed coral reefs in a cost-effective, non-invasive manner (\cite{rossi2020detecting18,zhong2023fine29,lange2020quick115}). Furthermore, these techniques support the production of realistic and richly textured orthomosaics, digital surface models. Integrating them with other survey methods across various environments can enhance the accuracy and efficiency of 3D coral habitat reconstruction, leading to more comprehensive and detailed assessments of coral reef ecology and environmental changes.

Early studies made preliminary attempts to use traditional photogrammetric methods for reconstructing underwater coral reefs  (\cite{andono20123d116,drap2013automating117,guo2016accuracy9}). Some research further incorporated photogrammetry with an underwater geodetic network to establish a consistent and unambiguous reference frame, thereby enabling precise coral monitoring (\cite{nocerino2020coral138,carlot2020community139,neyer2018monitoring140,rossi2020detecting18,casella2017mapping4}). With the advancements in photogrammetric computer vision, the emergence of image-based 3D reconstruction techniques such as Structure-from-Motion (SfM) (\cite{schonberger2016structure20}) and Multi-View Stereo (MVS)  (\cite{schonberger2016pixelwise46}) has enabled the automation of the reconstruction process. The typical photogrammetric 3D reconstruction workflow comprises two key stages: camera pose estimation and dense surface reconstruction. Camera pose estimation, achieved through techniques like SfM, determines the position and orientation of the cameras in 3D space by performing tasks like feature extraction, feature matching, and bundle adjustment. Dense surface reconstruction, on the other hand, focuses on generating a detailed model of the scene, often trying to estimate the 3D coordinates of each pixel, with MVS being the current predominant approach. Researchers have applied these techniques to 3D reconstruction of underwater coral reefs, achieving impressive results with 3D measurements accurate to the centimeter or even millimeter level  (\cite{zhong2023fine29,kalacska2018freshwater118,mohamed2020towards119}). However, these methods have limitations and may experience degraded performance or even failure under suboptimal conditions. Specifically, issues arise when images exhibit sparse or chaotic textures, high noise levels, occlusions, or insufficient overlap between images. Given the variable imaging conditions underwater and the complex, intricate structures of coral reefs, such as their tentacles, these challenging scenarios frequently arise, posing substantial demands on the accuracy, robustness, and efficiency of underwater 3D reconstruction technologies.

In recent years, researchers have not only advanced existing 3D reconstruction methodologies but have also turned their attention to rapidly advancing photogrammetric computer vision and deep learning techniques in pursuit of improved reconstruction effects. Over the past decade, advancements in camera pose estimation have included deep learning-based feature extraction methods (\cite{detone2018superpoint8,revaud2019r2d2_17,tyszkiewicz2020disk24,zhao2022alike64,edstedt2024dedode50,edstedt2024dedode51}), feature matching methods  (\cite{sarlin2020superglue19,sun2021loftr21,lindenberger2023lightglue41}), and end-to-end SfM frameworks  (\cite{wang2023visual90}). In dense surface reconstruction, traditional MVS methods have been enhanced by deep learning technologies, leading to the emergence of deep learning-based MVS methods  (\cite{zhang2023vis27,cao2024mvsformer++52}). Additionally, innovations in computer graphics, such as Neural Radiance Fields (NeRF) (\cite{mildenhall2021nerf14}) and Gaussian Splatting (GS)  (\cite{kerbl20233d31}), have introduced innovative solutions for dense surface reconstruction, fostering NeRF-based methods  (\cite{muller2022instant15,tancik2023nerfstudio22,li2023neuralangelo12}) and GS-based methods  (\cite{guedon2024sugar30,huang20242d32,yu2024gaussian86}). These cutting-edge solutions have demonstrated exceptional performance in certain application scenarios.

Despite the impressive performance of these advanced methods in testing environments, their applicability in underwater environments, particularly in coral reef scenes, remains to be further validated. Can these methods truly outperform traditional approaches when applied to coral reef images? While several studies have investigated their potential for high-quality coral reef modeling (\cite{zhong2023fine29,zhong2025high145,zhong2024application148}), a thorough and methodical comparison across methods is still lacking. Addressing this gap is essential for practical applications. Yet, without a comprehensive review and evaluation of both classical and cutting edge methods, researchers and practitioners lack the necessary evidence to select the most effective workflows. This absence of guidance hinders efforts to deploy scalable, reliable surveying solutions at a time when coral ecosystems are under unprecedented threat. This study therefore aims to:
\begin{itemize}
	\item systematically review the image based 3D reconstruction algorithms, with a focus on recent technological advancements, and provide a structured synthesis of their development and capabilities.
	\item benchmark the performance of various methods on both real and synthetic coral reef datasets, using a comprehensive set of evaluation metrics.
	\item provide evidence based recommendations for selecting the optimal solutions of coral reef modeling and discuss current limitations and identify promising directions for future research.
\end{itemize}
By bridging the divide between photogrammetric computer vision innovations and real world reef applications, our work delivers actionable guidance to support urgent conservation efforts with reliable, high fidelity 3D reconstruction solutions.

The rest of this paper is organized as follows: Section 2 first presents a typical 3D reconstruction workflow for underwater coral images, and then briefly reviews the technical developments addressing specific challenges, including comparative analyses of various methods. Section 3 provides a series of experimental evaluations of the 3D reconstruction solutions discussed in Section 2, performed using various coral reef image datasets. Section 4 further discusses the results presented in Section 3 and explores current challenges and potential future research directions. Section 5 concludes with key findings and their broader implications.


\section{3D Reconstruction Workflow and Method Evolution}\label{}

\begin{figure*}
	\centering
	\includegraphics[width=2\columnwidth]{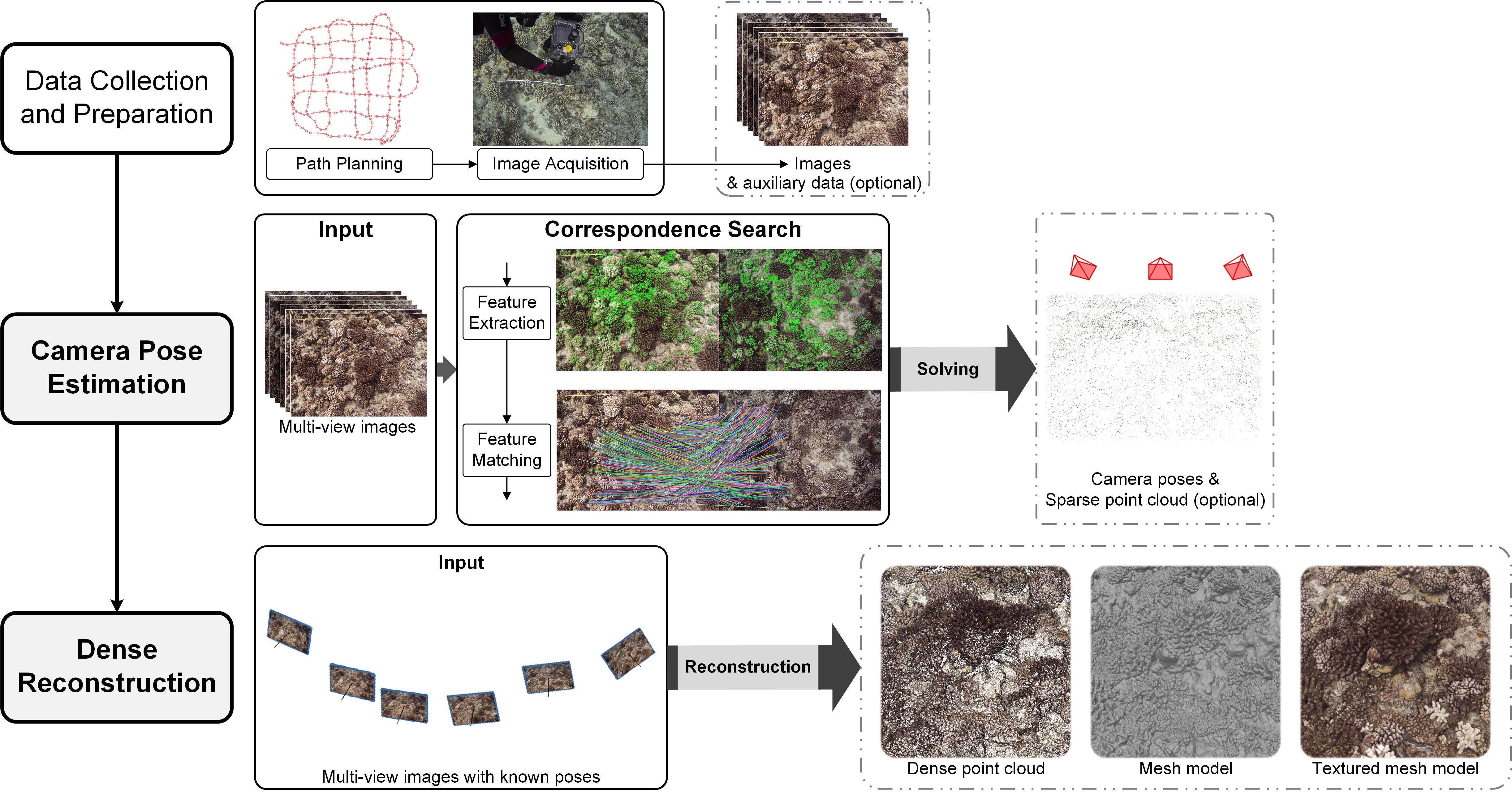}
	\caption{The pipeline for 3D surface reconstruction of coral reefs based on images.}
	\label{fig1}
\end{figure*}

The overall workflow for 3D reconstruction of coral reefs is illustrated in Figure \ref{fig1}. The process can be divided into three stages: data collection and preparation, camera pose estimation, dense surface reconstruction. The first stage involves acquiring underwater imagery and auxiliary data. Subsequently, camera poses are accurately estimated through the analysis of correspondences across multiple images. The dense surface reconstruction phase then provides detailed geometric information of the coral reefs. Finally, these reconstruction results are used to create products such as digital surface models, orthomosaics, and digital twins. The quality of both camera pose estimation and dense surface reconstruction is critical, as it directly influences the accuracy and reliability of the final photogrammetric products, making these stages pivotal to the workflow. In recent years, new techniques for these stages have developed rapidly, and their potential is a main focus of this paper.

\subsection{Data collection and preparation}

In the first stage, the primary objective is to acquire underwater images using visual sensors such as Digital Single Lens Reflex (DSLR) cameras and action cameras. The captured images should be high-resolution, exhibit sufficient overlap, and collectively cover the entire survey area to enable the application of photogrammetric computer vision techniques for 3D reconstruction of coral reef structures. The spatial resolution of the imagery directly determines the achievable reconstruction accuracy. For example, when monitoring subtle changes in coral morphology, which may be limited to only a few centimeters over the course of a year, it is preferable to achieve millimeter-level resolution. In practice, data collection is typically performed by divers or unmanned vehicles such as Remotely Operated Vehicles (ROVs) or Autonomous Underwater Vehicles (AUVs).


In addition to the image data, auxiliary measurements can be obtained using various instruments. For instance, total stations or GNSS technology can be employed to set up underwater control points for geo-referencing  (\cite{zhong2023combining53,jaud2023low56}). Inertial Measurement Units (IMUs) can provide approximate camera positions and orientations during imaging (\cite{nocerino2023camera54}). Laser or sonar can inherently deliver geometrical information with metric scale (\cite{istenivc2020automatic55,rahman2022svin2_60}). Additionally, calibration tools such as chessboards and color calibration boards are used for camera calibration (\cite{cahyono2020underwater57,skinner2017automatic58}). After data collection, preparation and pre-processing are necessary. Specifically, when color distortion occurs in images from relatively deep areas, radiometric correction should be applied to improve image quality and facilitate subsequent processing (\cite{neyer2019image59}).

\subsection{Camera pose estimation}\label{sparse}

Camera pose estimation focuses on determining the accurate positions and orientations of the cameras. Traditional aerial triangulation typically relies on initial estimates of camera poses to achieve accurate results, which can be a limitation in underwater environments that often lack reliable navigation data. In contrast, SfM automatically estimates camera poses without prior information, making it a powerful solution. Its speed, low cost, simplicity, and versatility have led to widespread adoption in 3D reconstruction tasks.

As illustrated in Figure \ref{fig1}, SfM is primarily composed of two main components: correspondence search and reconstruction. The first component aims to detect regions of overlap between input images and match corresponding projections of the same 3D points across overlapping areas. This process involves two critical steps: feature extraction and feature matching. Feature extraction identifies distinctive points (keypoints) in images along with their descriptors, which characterize the surrounding area of each keypoint. These features are extracted using local feature methods designed to be invariant to changes in translation, rotation, scale, and illumination. Subsequently, the matching process seeks to identify corresponding projections of the same points across overlapping images, thereby establishing correspondences between image pairs. Feature matching commonly involves geometrical verification to eliminate potential outliers in the matches (\cite{zhong2023deep61}). This verification is typically based on local photometric and geometric constraints. The quality of the correspondence search directly impacts the accuracy and reliability of the reconstruction. Existing solutions may face significant challenges in complex application scenarios, highlighting this as a key research problem.

The second component aims to estimate the image poses and 3D scene points using the correspondences between image pairs, which is achieved through image registration and triangulation in SfM reconstruction. The critical processing step in this phase is bundle adjustment (\cite{triggs2000bundle23}). Given that the extracted feature locations and correspondences may contain errors, and lens distortions can occur, errors can accumulate rapidly, potentially leading to drift in SfM. Bundle adjustment addresses these issues by minimizing the reprojection error—the discrepancy between the projected positions of 3D points in images and the actual locations of detected feature points—thereby improving the accuracy and consistency of the reconstruction. The objective function of bundle adjustment is presented as follow:
\begin{equation}\label{eq1}
	$$
	\underset{P_j,X_i}{\min}\sum^n_{i=1}\sum^m_{j=1}\rho_{ij}\left(\left\|{\pi (P_j,X_i)-x_j} \right\|^2_2\right),
	$$
\end{equation}

where $P_j$ represents the pose of an image, $X_i$ denotes a 3D scene point, $\pi(\cdot)$ is the function that projects the 3D point onto the image, $x_j$ indicates the location of a feature on the image, and $\rho_{ij}(\cdot)$ is a loss function used to potentially down-weight outliers. The notation $\|\cdot\|_2$ represents the L2 norm. This optimization process effectively reduces and corrects issues related to noise, mismatches, and initial estimation errors, resulting in refined camera poses and an accurate reconstructed scene structure.

Given the established framework for SfM processes, including pose estimation, triangulation, and bundle adjustment, the correspondence search emerges as the primary challenge, particularly in underwater coral environments. This process becomes unreliable in scenes with chaotic or sparse textures, or under significant variations in viewpoint or lighting conditions, which can hinder SfM reconstruction. Due to the greater impact of the water medium on imaging compared to the air medium, coral reef images captured underwater face challenges such as image degradation, similarity of natural textures, and cluttered background textures. These underwater-specific factors make correspondence search more challenging than in terrestrial applications. Based on a flexible framework for incremental SfM illustrated in Figure 2, the following subsections systematically review three critical technologies within this pipeline: feature extraction, feature matching, and reconstruction.

\begin{figure*}
	\centering
	\includegraphics[width=1.6\columnwidth]{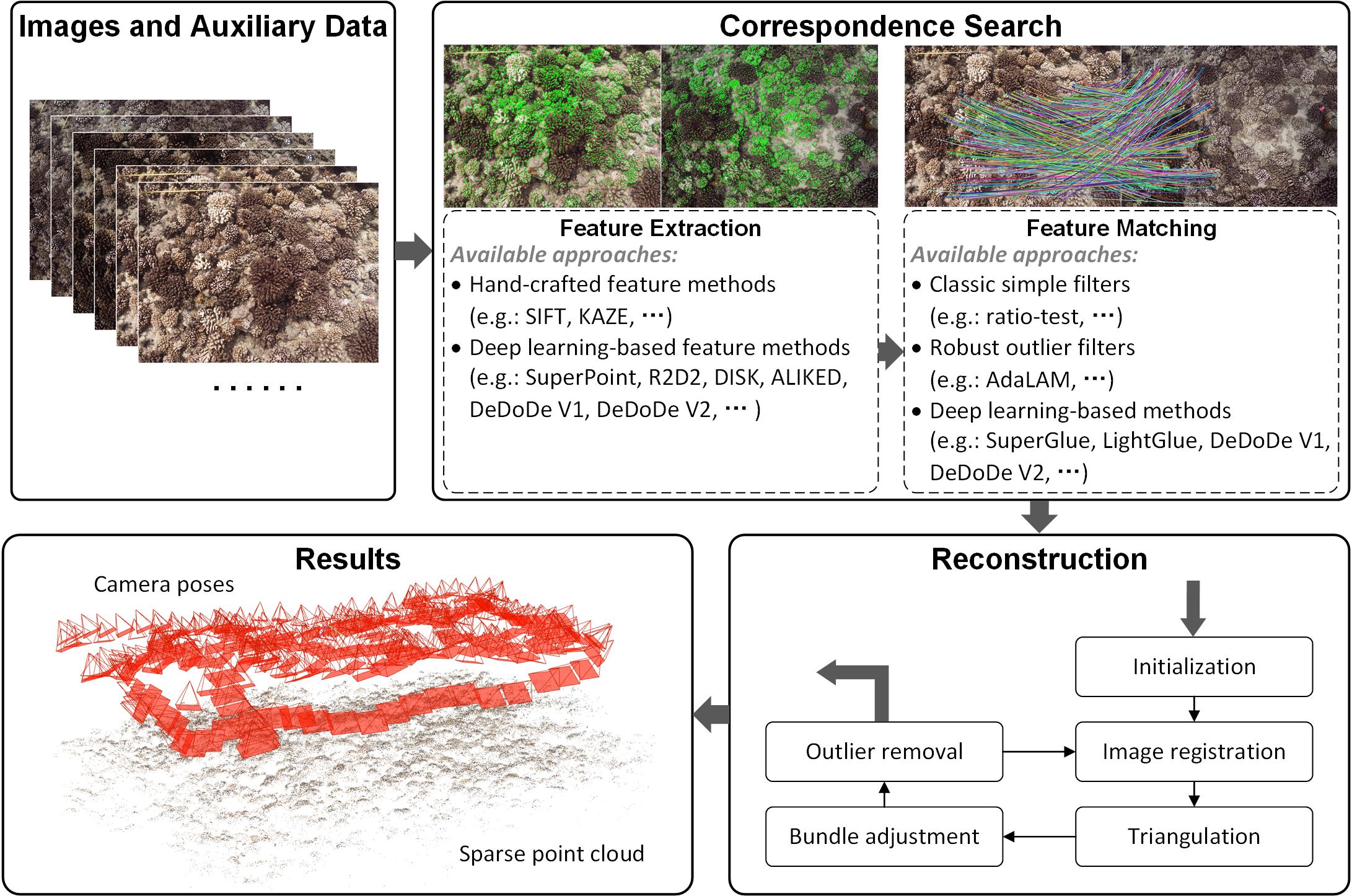}
	\caption{A flexible framework for incremental Structure-from-Motion.}
	\label{fig2}
\end{figure*}

\subsubsection{Feature extraction}

SfM relies on local features to align overlapping images by identifying the same scene points. These features must be invariant to radiometric and geometric changes to ensure accurate correspondence between different images. Traditional hand-crafted feature extraction methods typically follow a two-stage pipeline: keypoint detection followed by descriptor computing. The geometric invariance of these descriptors primarily addresses scale invariance and orientation invariance, which have been extensively studied over the past decades. SIFT  (\cite{lowe1999object93,lowe2004distinctive13}) is widely known for its scale and orientation invariance by estimating keypoint scales using gradient histograms. SURF (\cite{bay2006surf35}) significantly improves the computational efficiency of SIFT through various optimizations, while ORB features  (\cite{rublee2011orb37}) further simplify and accelerate the process. Alcantarilla et al. introduced KAZE features  (\cite{alcantarilla2012kaze1}), which leverage non-linear scale space through non-linear diffusion filtering to achieve invariance to rotation, scale, and limited affine transformations. They also proposed an enhanced version, AKAZE  (\cite{alcantarilla2011fast63}), which improves processing speed. However, these methods are based on heuristics, which can be difficult to generalize across different scenarios, potentially leading to reconstruction failures. 

With the rapid advancement of deep learning, researchers have turned their attention to learning-based local feature methods, aiming to address the limitations of traditional hand-crafted approaches in extreme scenarios. Early learning-based local feature methods, such as LIFT  (\cite{yi2016lift26}), improved performance by integrating region detector, orientation estimator and feature descriptor in a single differentiable network. DeTone et al. proposed an end-to-end self-supervised local feature method called SuperPoint  (\cite{detone2018superpoint8}), which trains the network on homography image pairs. Additionally, many methods integrate keypoint detection into the pipeline by jointly training feature detection and description. For example, Revaud et al. proposed R2D2  (\cite{revaud2019r2d2_17}), which employs a Siamese decoding structure to achieve reliable keypoint detection and description. Tyszkiewicz et al. introduced DISK  (\cite{tyszkiewicz2020disk24}), which uses a joint training objective for keypoint detection and description through reinforcement learning. Zhao et al. introduced ALIKE  (\cite{zhao2022alike64}), which features a Differentiable Keypoint Detection module designed for keypoint training and achieves high efficiency through a lightweight design. Later, they incorporated deformable convolutions and developed ALIKED  (\cite{zhao2023aliked28}), resulting in improvements in both performance and efficiency. In contrast, Edstedt et al. developed DeDoDe  (\cite{edstedt2024dedode50}) and DeDoDe V2  (\cite{edstedt2024dedode51}), which decouple the detection and description steps into two independent models, optimizing each separately to enhance robustness and accuracy. The research above illustrates that deep learning-based local feature methods offer significant advantages over traditional hand-crafted methods in various aspects, thereby facilitating the acquisition of more and higher-quality correspondences in practical applications, which benefits SfM reconstruction. A comparison of these methods is presented in Table \ref{tab1}.

\begin{table*}[ht]
	\caption{Comparison of representative local feature extraction methods (PY, Publication year).}
	\centering
	\label{tab1}
	\renewcommand\arraystretch{1.1}
	\begin{tabular}{lllll}
		\hline
		Category  & Method     & PY   & Benefits and obstacles     &  \\
		\hline
		\multirow{5}{*}{Hand-crafted methods}        & SIFT       & 1999 & \multirow{5}{*}{\begin{tabular}{p{9cm}} \textbf{Advantages}: (1) Rapid computational performance; (2) Simple   and comprehensible principles; (3) High accuracy in keypoint localization.\\ \textbf{Disadvantages}: (1) Significant sensitivity to parameter settings;   (2) Performance instability under extreme or challenging conditions.\end{tabular}} &  \\
		& SURF       & 2008 &  &  \\
		& ORB        & 2011 &  &  \\
		& KAZE       & 2012 & &  \\
		& AKAZE      & 2013 & &  \\
		\hline
		\multirow{8}{*}{Deep learning-based methods} & LIFT       & 2016 & \multirow{8}{*}{\begin{tabular}{p{9cm}} \textbf{Advantages}: (1) Enhanced descriptive capability and   robustness; (2) Simple parameter adjustment.\\  \textbf{Disadvantages}: (1) Greater complexity of the model architecture;   (2) Dependence on the datasets; (3) Require more computational power.\end{tabular}}                                      &  \\
		& SuperPoint & 2018 & &  \\
		& R2D2       & 2019 &   &  \\
		& DISK       & 2020 &  &  \\
		& ALIKE      & 2022 &   &  \\
		& ALIKED     & 2023 &        &  \\
		& DeDoDe     & 2024 & &  \\
		& DeDoDe V2  & 2024 &    & \\ 
		\hline
	\end{tabular}
\end{table*}

\subsubsection{Feature matching}

After feature extraction, a set of keypoints and their corresponding descriptors are obtained. For SfM reconstruction, accurate feature matching is required to establish correspondences between images of the same scene region. The problem is classically solved by matching a keypoint with its most similar counterpart in another image, specifically identifying the nearest neighbor in descriptor space. However, under conditions such as low image overlap, sparse or overly similar textures, and variations in lighting, descriptors may fail to accurately represent features. This can result in many outliers—matches that are incorrect or significantly different from the expected correspondences. The classical ratio-test  (\cite{lowe2004distinctive13}) is a basic and straightforward feature matching method, often used in conjunction with mutual nearest neighbor checks for rapid correspondence search. While these simple filters are efficient and widely applied, their performance is limited. They often miss many outliers or incorrectly filter out inliers, which can complicate SfM solutions. To achieve more robust and reliable feature matching, researchers have explored techniques such as local spatial consistency checks and global geometric verification. Among these methods, one of the most representative is AdaLAM (\cite{cavalli2020adalam5}), proposed by Cavalli et al. AdaLAM demonstrates significant robustness and is notable for its ease of deployment and high operational efficiency. While AdaLAM integrates several ideas from previous research, it remains based on geometrical assumptions, such as local affine consistency, which may limit its applicability in certain scenarios—especially when dealing with no-planar surfaces. To address this, it employs adaptive strategies to relax these assumptions, enhancing its generalization across different domains. Compared to general methods, AdaLAM not only significantly reduces outliers but also increases the number of inliers.

Like for feature extraction, recent advancements have introduced deep learning-based approaches that aim to overcome the limitations of manual descriptor engineering and enhance generalization capabilities. Among these, SuperGlue (\cite{sarlin2020superglue19}) is a significant advancement that uses a graph neural network framework to effectively match extracted feature points and address mismatches. It also leverages optimal transport  (\cite{peyre2019computational68}) combined with Transformers  (\cite{vaswani2017attention67}) to resolve the partial assignment problem. The architecture of SuperGlue includes an attentional graph neural network and an optimal matching layer, utilizing self-attention for individual images and cross-attention for image pairs, enabling high-quality feature matching. LightGlue  (\cite{lindenberger2023lightglue41}) offers several improvements, notably its ability to introspect the confidence of its own predictions. Compared to SuperGlue, LightGlue enhances accuracy, efficiency, and training complexity.

In recent years, novel detector-free local feature matching methods have emerged, such as LoFTR  (\cite{sun2021loftr21}) and ASpanFormer  (\cite{chen2022aspanformer65}). These approaches focus on generating correspondences directly from image pairs, circumventing the traditional process of keypoint extraction and description. He et al. developed a framework known as Detector-free SfM (DF-SfM)  (\cite{he2024detector66}) for these detector-free methods, which has exhibited good performance in scenarios with sparse texture. However, they are constrained by the resolution of input images, which can adversely impact the spatial accuracy of correspondences. 

\subsubsection{SfM reconstruction}

SfM methods can be categorized into two main categories based on their image processing strategies: incremental SfM and global SfM. Incremental SfM (\cite{schonberger2016structure20}) follows a step-by-step approach, progressively adding images and iteratively refining camera poses and 3D scene points through local and global optimizations. This allows for the refinement of estimates as new images are incorporated, which is particularly beneficial because good initial values are critical for effective adjustment optimization. In contrast, global SfM (\cite{pan2024global95}) processes all images simultaneously. This approach involves initial global estimation of camera poses, followed by comprehensive global optimization and subsequent triangulation. While this method offers a holistic view of the scene, it relies heavily on the accuracy of the initial estimates. Due to its iterative refinement, incremental SfM has become the dominant approach in contemporary applications (\cite{jiang2020efficient34}). As shown in Figure \ref{fig2}, the reconstruction process typically begins by selecting an image pair with sufficient overlap and correspondences to estimate initial poses and 3D points. Subsequently, additional images are registered to the current model and combined with already registered images for triangulation. Bundle adjustment (\cite{triggs2000bundle23}) is performed to mitigate the effects of cumulative errors and exclude unreliable observations. This iterative procedure allows for the accurate estimation of all camera poses.

The aforementioned typical SfM methods can be referred to as feature correspondence-based SfM and have been extensively studied, with numerous classical solutions emerging, such as COLMAP (\cite{schonberger2016structure20}). Recently, end-to-end SfM methods have emerged to circumvent explicit feature matching by directly regressing camera poses (\cite{zhou2017unsupervised69,vijayanarasimhan2017sfm70}) or employing differential bundle adjustment, where a feed-forward multilayer perceptron is trained to predict the damping factor during optimization (\cite{tang2018ba71}). This strategy helps avoid the issues associated with low-quality correspondences. Some techniques leverage advanced deep 2D point tracking to extract pixel-accurate tracks and utilize differentiable components for end-to-end training (\cite{wang2023visual90}). Despite these advancements, such methods currently face limitations in scalability in real-world applications and are still in the exploratory phase. Furthermore, while deep learning-based multi-view refinement have shown promise in improving correspondence accuracy (\cite{he2024detector66,lindenberger2021pixel72}), it has not yet generalized well across diverse scenarios. 

\subsection{Dense surface reconstruction}\label{dense}


Although SfM reconstruction produces 3D point clouds through triangulation, the resulting point clouds are typically too sparse to represent the detailed geometric information of the observed coral reef areas. This limitation is particularly pronounced in underwater coral reef environments, where fine-scale structures such as small coral branches or tentacles are of ecological significance. To address this, dense surface reconstruction techniques are employed to recover high-resolution representations of the scene. This limitation necessitates the use of dense surface reconstruction techniques to recover the finer morphological details essential for accurate monitoring and analysis.

Dense surface reconstruction refers to the process of estimating 3D surface geometry by integrating information across multiple image views. Contemporary methods for dense reconstruction employ a range of approaches, but they all share a common principle of generating a dense representation of the 3D scene, such as a dense point cloud or mesh. This process depends on the geometric relationships and viewpoint consistency among multi-view images for inference and optimization. Consequently, it necessitates the intrinsic and extrinsic camera parameters, which can be derived from SfM reconstruction. The quality of the dense surface reconstruction directly influences downstream tasks such as measurement, scene analysis, and digital modeling.

\begin{figure*}
	\centering
	\includegraphics[width=2.0\columnwidth]{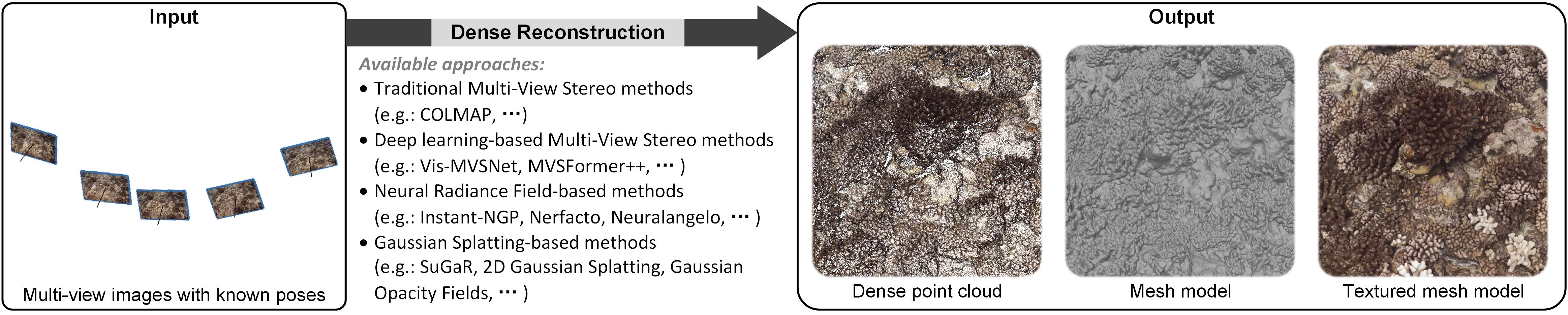}
	\caption{A variety of dense surface reconstruction technologies applicable to coral reefs.}
	\label{fig3}
\end{figure*}

In the context of coral reef observation, dense surface reconstruction offers the ability to model fine morphological details, facilitating precise structural analysis. However, coral reefs present unique challenges due to their complex and diverse forms. Features such as densely packed coral polyps, overhangs, and branched structures often lead to occlusions and self-similarity, which can confound reconstruction algorithms. Environmental factors such as uneven lighting, water turbidity, and suspended particles further degrade image quality, introducing noise and reducing feature consistency across views. Moreover, structural changes in coral reefs can occur on the scale of mere centimeters over annual timescales, demanding reconstruction methods with exceptionally high spatial accuracy. Given these challenges, there is an urgent need for high-quality and efficient dense reconstruction solutions. As illustrated in Figure \ref{fig3}, current dense surface reconstruction methods can be broadly categorized into four categories: traditional multi-view stereo (MVS) methods, deep learning-based MVS methods, methods based on neural radiance fields (NeRFs), and methods based on Gaussian Splatting (GS). Each category offers distinct advantages and limitations in terms of reconstruction quality and computational efficiency. Representative approaches within these categories are summarized in Table \ref{tab2} and will be discussed in detail in the following subsections. 

\begin{table*}[ht]
	\caption{Comparison of representative dense reconstruction methods (PY, Publication year).}
	\centering
	\label{tab2}
	\renewcommand\arraystretch{1.1}
	\begin{tabular}{lllll}\hline
		Category    & Method                  & PY   & Type of   output  & Benefits   and obstacles  \\ 
		\hline
		\begin{tabular}{p{2.2cm}}{Traditional   MVS methods}\end{tabular}                           & COLMAP                  & 2016 & Point cloud       & \begin{tabular}{p{7.5cm}}\textbf{Advantages}: (1) Well-established theoretical foundations with clear operational principles.\\ \textbf{Disadvantages}: (1) Dependent on parameter settings; (2) Performance heavily depends on texture and lighting conditions.\end{tabular}                                                     \\
		\hline
		\begin{tabular}{p{2.2cm}}\multirow{6}{*}{\parbox{2.2cm}{Deep   learning-based MVS methods}}\end{tabular} & SurfaceNet              & 2017 & Point cloud       & \multirow{6}{*}{\begin{tabular}{p{7.5cm}}\textbf{Advantages}: (1) High adaptability; (2) Efficient in processing.\\ \textbf{Disadvantages}: (1) Dependent on training data; (2) Complexity of the learned features and their interaction with the MVS pipeline.\end{tabular}}                                                       \\
		& MVSNet                  & 2018 & Point   cloud     &  \\
		& PatchmatchNet           & 2021 & Point   cloud     &  \\
		& Vis-MVSNet              & 2022 & Point   cloud     &     \\
		& MVSFormer               & 2022 & Point   cloud     &    \\
		& MVSFormer++             & 2024 & Point   cloud     &     \\
		\hline
		\begin{tabular}{p{2.2cm}}\multirow{6}{*}{\parbox{2.2cm}{Methods based on NeRF}}\end{tabular}            & NeuS                    & 2021 & Mesh              & \multirow{6}{*}{\begin{tabular}{p{7.5cm}}\textbf{Advantages}: (1) Capable of achieving high-quality reconstruction; (2) Suitable for high-fidelity real-time rendering.\\  \textbf{Disadvantages}: (1) Long training times; (2) High computational resource demands; (3) Sensitivity to data and parameter settings.\end{tabular}} \\
		& Instant-NGP             & 2022 & Radiance   fields &   \\
		& Nerfacto                & 2023 & Radiance   fields &         \\
		& Neuralangelo            & 2023 & Mesh              &     \\
		& NeuS2                   & 2023 & Mesh              &       \\
		& BakedSDF                & 2023 & Mesh              &      \\
		\hline
		\begin{tabular}{p{2.2cm}}\multirow{3}{*}{\parbox{2.2cm}{Methods based on GS}}\end{tabular}               & SuGAR                   & 2023 & Mesh              & \multirow{3}{*}{\begin{tabular}{p{7.5cm}} \textbf{Advantages}: (1) Improved computational efficiency; (2) Capable of providing high-quality detail representation.\\ \textbf{Disadvantages}: (1) Immature in stability and robustness; (2) Sensitivity to data and parameter settings.\end{tabular}}               \\
		& 2D GS   & 2024 & Mesh              &    \\
		& GOF & 2024 & Mesh              &      \\
		&  &  &               &      \\
		\hline                                                                                                                                                                                                                                                                                                                      
	\end{tabular}
\end{table*}

\subsubsection{Traditional multi-view stereo methods}

Over the past decades, MVS methods have become one of the most widely applied techniques for dense reconstruction (\cite{liu2020depth74}), demonstrating significant potential for efficiently reconstructing intricate scenes. MVS techniques are based on the same principles as classic binocular stereo, where corresponding pixels between images are identified using manually designed visual similarity metrics. Typically, MVS requires considering correspondences between multiple neighboring images when estimating the depth of a reference image. This approach leverages redundant observations to achieve more precise and reliable depth estimates, making it essential to handle varying viewpoints effectively.

According to output representations, MVS algorithms can be categorized into three classes: direct point cloud reconstruction, volumetric reconstruction and depth map reconstruction (\cite{yao2018mvsnet25}). Among these, depth map-based methods are particularly favored for their flexibility and simplicity, making them well-suited for reconstructing large-scale 3D structures (\cite{liu2020depth74}). These methods use dense stereo matching techniques to generate depth maps for each reference image, utilizing several neighboring images. The depth maps are then fused to produce a dense 3D point cloud. To generate higher-quality depth maps, various solutions have been developed, with the PatchMatch algorithm (\cite{barnes2009patchmatch48}) being one of the most efficient and robust methods for MVS scenarios. This algorithm leverages the natural coherence of the images and effectively addresses fronto-parallel bias. Due to its performance, MVS methods based on PatchMatch have achieved top scores in benchmark challenges (\cite{shen2013accurate76,galliani2015massively75,schonberger2016pixelwise46}), leading to their widespread application in contemporary 3D reconstruction software.

\subsubsection{Deep learning-based multi-view stereo methods}

Driven by advancements in deep learning, learning-based MVS approaches have emerged. Unlike traditional methods, which address MVS problems through iterative propagation and matching processes, learning-based MVS methods utilize deep neural networks to achieve high-quality reconstruction in an end-to-end fashion. These methods demonstrate significant potential as alternatives to conventional MVS techniques, operating on the principle of feature matching along epipolar lines given known camera poses.

Hartmann et al. (\cite{hartmann2017learned82}) introduced a learnable multi-view cost metric to measure the multi-view photo-consistency between image patches. Subsequently, Ji et al. proposed SurfaceNet (\cite{ji2017surfacenet49}), which learns cost volume regularization from geometry ground truth. Since then, various networks have been developed, with MVSNet (\cite{yao2018mvsnet25}) being one of the most notable. MVSNet begins by extracting deep visual image features and then employs differentiable homography warping to build a 3D cost volume within the reference camera frustum. The method then applies regularization and refinement processes to produce the final output. MVSNet is also the first end-to-end method for learning depth map inference in MVS, and it has had a significant impact on subsequent approaches. Many later methods have built upon its framework. For example, Vis-MVSNet (\cite{zhang2023vis27}) extends MVSNet by formulating a more reliable cost volume to address the reconstruction of scenes with severe occlusions, achieving improved results. Some learning-based approaches have also sought to leverage the PatchMatch algorithm to avoid global cost volumes. Among these, PatchmatchNet (\cite{wang2021patchmatchnet81}) introduced the first end-to-end cascade formulation of PatchMatch. Later, inspired by the significant achievements of Vision Transformers (ViT) (\cite{dosovitskiy2020image83}) in various visual tasks, researchers have increasingly integrated transformers into MVS learning. MVSFormer (\cite{cao2022mvsformer80}) represents a pioneering effort in this area, combining pre-trained ViTs for feature extraction with integrated architectures and training strategies, and has demonstrated promising results. Furthermore, methods such as TransMVSNet (\cite{ding2022transmvsnet79}) and MVSFormer++ (\cite{cao2024mvsformer++52}) have been developed, with MVSFormer++ achieving state-of-the-art results on the indoor DTU dataset (\cite{aanaes2016large77}) and ranking top-1 on the outdoor Tanks-and-Temples dataset (\cite{knapitsch2017tanks78}).

\subsubsection{Neural radiance field-based methods}

Beyond MVS, research has also explored various alternative approaches for 3D reconstruction, with one notable category being NeRF-based methods. Building on the ground-breaking work of Mildenhall et al. (\cite{mildenhall2021nerf14}), NeRF technology has experienced rapid development and has been extensively applied in photo-realistic novel view synthesis and 3D scene representation. The original NeRF framework represents the 3D scene structure and appearance implicitly as a continuous 5D radiance field, encompassing both location and viewing direction, based on images with known poses. It samples 5D coordinates along camera rays and inputs these locations into a Multilayer Perceptron (MLP) to predict color and volume density. Subsequently, volume rendering techniques are employed to synthesize the image from these predictions. Given that the rendering function is differentiable, this scene representation can be optimized by minimizing the difference between the rendered and ground truth images through gradient-based methods.

Although NeRF was initially developed for novel view synthesis, some methods have leveraged the continuity inherent in MLPs and neural volume rendering to enable optimized surfaces to interpolate reasonably across spatial locations, thereby achieving smooth and complete surface representations (\cite{wang2021neus84,li2023neuralangelo12}). As a result, NeRF has become as a promising alternative to MVS for dense reconstruction. A representative example is NeuS (\cite{wang2021neus84}), which uses a neural network-encoded Signed Distance Field (SDF) for high-quality reconstruction of small objects, but its training process is time-consuming. Instant-NGP (\cite{muller2022instant15}) improves training efficiency by employing multi-resolution hash encoding to balance computational demand with accuracy, although it lacks surface constraints, resulting in significant noise in the geometry extracted from the learned density fields. Neus2 (\cite{wang2023neus2_85}) builds on the multi-resolution hash encoding and simplifies calculation of second-order derivatives to accelerate the process. Notably, Nerfacto (\cite{tancik2023nerfstudio22}) combines several advantageous features from prior methods, thereby achieving a balance between accuracy and efficiency. Furthermore, Nerfacto makes a significant contribution to the field by offering the Python framework Nerfstudio, which supports the export of results as point clouds and meshes. Neuralangelo (\cite{li2023neuralangelo12}) applies multi-resolution 3D hash grids and numerical gradients with coarse-to-fine optimization to achieve high-fidelity surface reconstruction. BakedSDF (\cite{yariv2023bakedsdf87}), on the other hand, initially optimizes a hybrid neural volume-surface scene representation and then converts it into a high-quality triangle mesh through pre-computation and transformation. 

\subsubsection{Gaussian Splatting-based methods}

The introduction of 3D Gaussian Splatting (\cite{kerbl20233d31}) in 2023 has presented new opportunities for dense surface reconstruction. This technique represents a scene using a set of 3D Gaussians, enabling photorealistic novel view synthesis while ensuring efficient training and real-time rendering. Specifically, for a given set of training images, the original method initializes the Gaussians using a sparse point cloud generated by SfM reconstruction. It then optimizes the positions, orientations, appearances, and alpha blending of numerous small 3D Gaussians to align the renderings as closely as possible with the input images. Gaussians are added or removed during the optimization process to better fit the scene’s geometry. This approach can capture the scene's geometric structure and appearance effectively, with high-speed rendering using a rasterizer.

While Gaussians can achieve high-quality scene rendering, their application for surface extraction presents significant challenges and limitations that impact the accuracy and reliability of reconstruction. Issues include the unstructured nature of optimized Gaussians, multi-view inconsistencies, and conflicts between volumetric 3D Gaussians and the thin nature of surfaces (\cite{guedon2024sugar30,huang20242d32,yu2024gaussian86}). To address these challenges, researchers have proposed various solutions. SuGAR (\cite{guedon2024sugar30}) introduces a regularization term to better align Gaussians with the scene's surface, and employs Poisson reconstruction to extract a mesh from the Gaussians, enabling accurate estimation of fine geometric details while being fast and user-friendly. NeuSG (\cite{chen2023neusg89}) utilizes normal priors predicted by neural implicit models to refine the point cloud generated by 3D Gaussian Splatting for more accurate surface reconstruction. Huang et al. proposed 2D Gaussian Splatting (2D GS) (\cite{huang20242d32}), which uses 2D Gaussians for a more precise representation of the scene and employs TSDF fusion to reconstruct the mesh. Gaussian Opacity Fields (GOF)(\cite{yu2024gaussian86}) , derived from ray-tracing-based volume rendering of 3D Gaussians, directly extracts geometry from 3D Gaussians, eliminating the need for Poisson reconstruction or TSDF fusion.

\section{Experiments}\label{}

\subsection{Experimental datasets}

\begin{figure*}
	\centering
	\includegraphics[width=2.0\columnwidth]{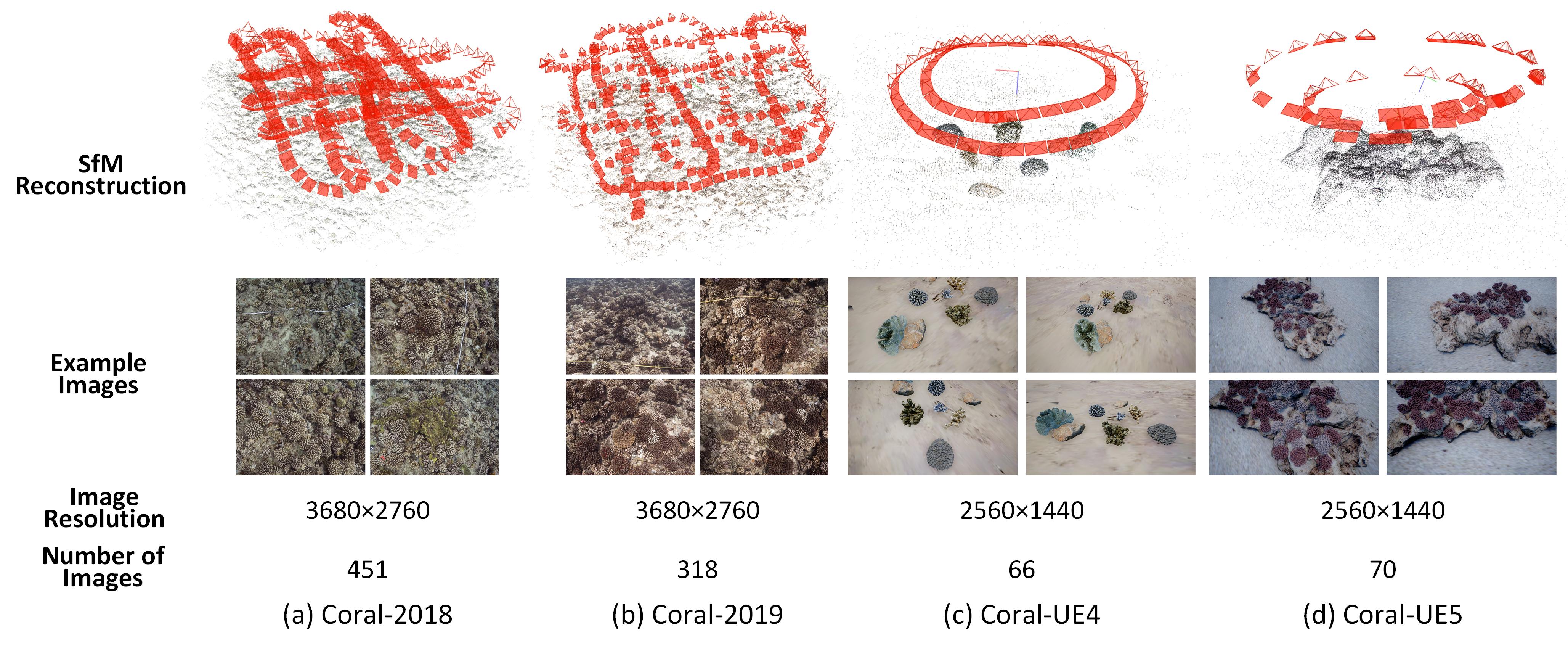}
	\caption{The overview of experimental datasets.}
	\label{fig4}
\end{figure*}

This section will qualitatively and quantitatively evaluate the current camera pose estimation and dense surface reconstruction solutions. The data used in the experiments can be categorized into two types: real-world images and synthetic images (or simulated images). Figure \ref{fig4} provides an overview of the basic information and example images for each dataset.

The real-world image data is provided by the Moorea Island Digital Ecosystem Avatar (IDEA) project. These images were captured underwater using a digital camera system that includes a PANASONIC LUMIX GH5S camera body, with a resolution of 3680×2760, and a wide-angle Lumix G 14mm f/2.5 lens. In this study, the data is organized into two datasets, named Coral-2018 and Coral-2019. The Coral-2018 dataset was collected in August 2018 and consists of 451 images, while the Coral-2019 dataset was collected in August 2019 and contains 318 images. All images in each dataset were taken along pre-planned paths, with an overlap rate between adjacent images ranging from approximately 70\% to 85\%, making them suitable for high-quality multi-view 3D reconstruction.

The synthetic datasets utilized in this study are created using the AirSim platform (\cite{shah2018airsim92}), a simulator built on Unreal Engine (\cite{sanders2016introduction91}) that offers physically and visually realistic simulations. We imported models of coral reefs, rocks, and other elements into the simulation environments. Then, we applied the camera functionality to capture images of the coral reefs from various viewpoints. This approach yielded high-resolution, distortion-free images with pose ground truth, suitable for assessment purposes. Two synthetic datasets are created for this study: Coral-UE4 and Coral-UE5, based on Unreal Engine 4 and Unreal Engine 5, respectively. In Coral-UE4, the scenes are rendered without the inclusion of additional participating media, resulting in clean and unperturbed imaging conditions. In contrast, an underwater blueprint is used in Coral-UE5 to simulate the underwater environment. This system enables the inclusion of volumetric water effects such as light scattering, absorption, and attenuation, allowing for a more realistic approximation of the optical properties of water and their impact on image formation. The images in both datasets have a resolution of 2560×1440 pixels and were captured from a 360-degree perspective around the scenes. The Coral-UE4 dataset features simpler scenes with less texture, while the Coral-UE5 dataset includes more complex and richly textured environments. To evaluate the performance of different solutions in more challenging scenarios, both datasets are downsampled by uniformly sampling one-fourth of the images from  the original dataset, resulting in lite datasets: Coral-UE4 (lite) with 17 images and Coral-UE5 (lite) with 18 images. The datasets and projects from this study are available at https://github.com/Atypical-Programmer/Coral3D.

\subsection{Camera pose estimation results comparison}\label{E2}

\subsubsection{Feature extraction and matching evaluation}\label{E2S1}

Here we first evaluate their applications in camera pose estimation. For feature extraction, we consider two hand-crafted methods, SIFT (\cite{lowe2004distinctive13}) and KAZE (\cite{alcantarilla2012kaze1}), as well as five deep learning-based approaches: SuperPoint (\cite{detone2018superpoint8}), R2D2 (\cite{revaud2019r2d2_17}), DISK (\cite{tyszkiewicz2020disk24}), ALIKED (\cite{zhao2022alike64}), and DeDoDe (\cite{edstedt2024dedode50}). It should be noted that DeDoDe V2 (\cite{edstedt2024dedode51}) has not yet been fully open sourced. The number of features extracted by each method is limited to 8000. For feature matching, two non-learning methods are employed: ratio-test (\cite{lowe2004distinctive13}) and AdaLAM (\cite{cavalli2020adalam5}). Additionally, three deep learning-based methods are used: SuperGlue (\cite{sarlin2020superglue19}), LightGlue (\cite{lindenberger2023lightglue41}), and LoFTR (\cite{sun2021loftr21}). Notably, LoFTR is a detector-free feature matching method that does not require keypoints. SuperGlue is available with two sets of pre-trained weights: one trained on the ScanNet dataset (\cite{dai2017scannet96}) for indoor scenes, referred to as SG (indoor), and another trained on the MegaDepth dataset (\cite{li2018megadepth97}) for outdoor scenes, referred to as SG (outdoor). Since the data augmentation during the training process of SuperGlue incorporates SuperPoint, this method exhibits optimal performance when paired with SuperPoint. Similarly, pre-trained weights for LightGlue are available with SIFT, SuperPoint, DISK, and ALIKED features, all of which are included in the comparative analysis. LoFTR is implemented using the Python library “kornia” and provides three sets of pre-trained weights for correspondence search, i.e. “indoor model”, “new indoor model”, and “outdoor model”.

\begin{figure*}
	\centering
	\includegraphics[width=2.0\columnwidth]{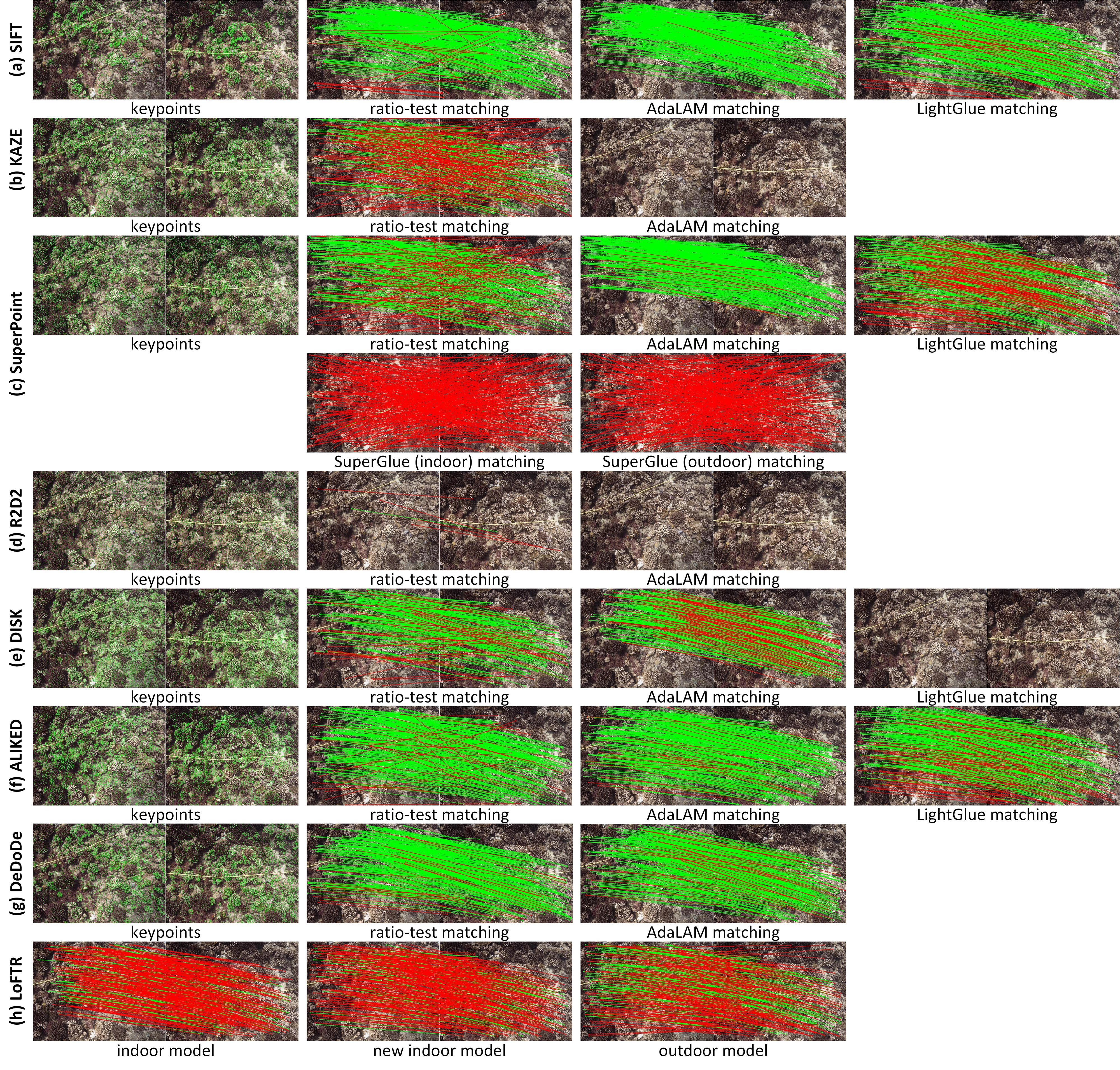}
	\caption{Visualization of the matching results of coral reef images with rotation and translation. The correct matches are depicted with green lines, while mismatches are represented with red lines. To ensure clarity, up to 400 matches are randomly selected from each result for display.}
	\label{fig5}
\end{figure*}

Preliminary experimental analysis reveals that while all methods effectively handle image pairs with high overlap and minimal color and viewpoint variations, significant differences emerge when dealing with more challenging data. Figure \ref{fig5} illustrates the feature extraction and matching results of various methods on an image pair characterized by rich textures, overlapping regions, and transformations such as translation and rotation. To better assess the performance of each method, we manually annotate correspondences to calculate the fundamental matrix and epipolar lines between the two images. After matching features, we calculate the distance between each feature in the reference image and its corresponding feature in the query image relative to the epipolar line, considering a match correct if the distance does not exceed 5 pixels.

Our experiments reveal distinct behaviors between hand-crafted and deep learning-based feature extraction and matching methods. Hand-crafted methods like SIFT and KAZE focus on regions with rich textures and lighting, such as tentacles and edges of light-colored corals. However, KAZE features often produce mismatches due to sensitivity to repetitive textures and noise, making them less reliable. Deep learning-based methods, particularly SuperPoint and DISK, provide a more uniform and dense distribution of keypoints. Among matching methods, the ratio-test generates many correct matches but also introduces errors, especially with KAZE. AdaLAM significantly reduces mismatches with SIFT, SuperPoint, and ALIKED, but fails with KAZE, indicating incompatibility with its descriptors. DeDoDe's larger descriptor proves more robust, outperforming its smaller counterpart. LightGlue is the most effective deep learning-based matcher, delivering accurate matches and supporting stable SfM reconstruction, while SuperGlue struggles with a high rate of mismatches. LoFTR's outdoor model performs best in coral reef environments, suggesting its alignment with such conditions over other models. Overall, LightGlue and AdaLAM provide superior matching performance, contributing to more accurate and stable reconstructions.

\begin{figure}
	\centering
	\includegraphics[width=1.0\columnwidth]{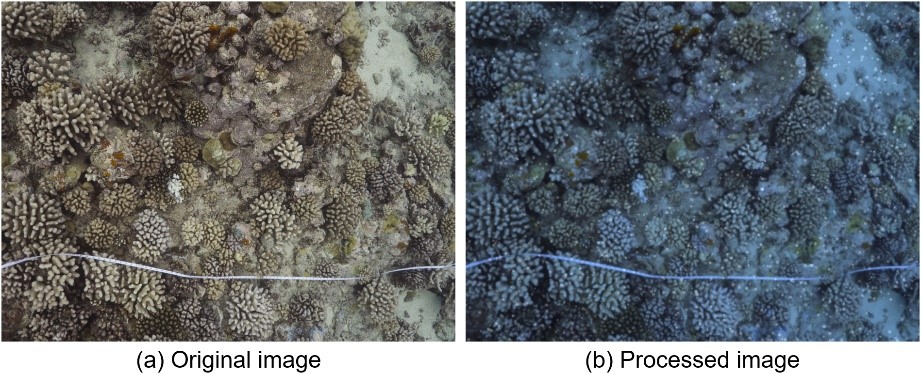}
	\caption{Original image and simulated image in the underwater conditions.}
	\label{fig6}
\end{figure}

\begin{figure*}
	\centering
	\includegraphics[width=2.0\columnwidth]{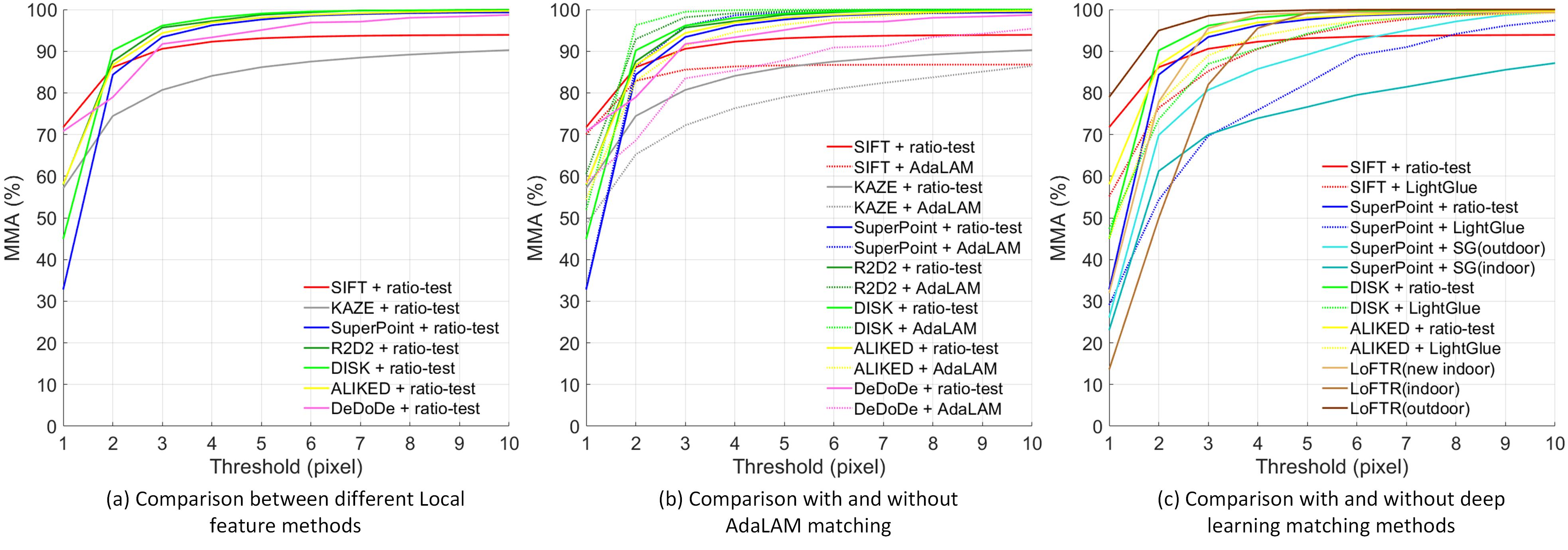}
	\caption{Mean Matching Accuracy (MMA) under simulated underwater conditions.}
	\label{fig7}
\end{figure*}

To quantitatively assess the ability of various methods to handle factors such as illumination changes, rotations, blurring, color shifts, and noise, we randomly select several hundred coral images to construct synthetic datasets with ground truth. To simulate the effects of underwater color shifts, suspended particles, and dispersion, we reduce the image brightness, enhance the blue and green channels, randomly add particles of varying sizes, and apply Gaussian filtering. An example of the resulting image pair is shown in Figure \ref{fig6}. We then perform correspondence searches on the images before and after processing using each solution and calculate the Mean Matching Accuracy (MMA), with the results presented in Figure \ref{fig7}. MMA is calculated by measuring the percentage of correct matches within a certain pixel error threshold for each image pair, and then averaging this accuracy across all image pairs. If no matches are found, the accuracy will be recorded as 0.

For the local feature extraction methods combined with the ratio-test, as shown in Figure \ref{fig7}(a), SIFT demonstrates the highest accuracy at a 1-pixel threshold, likely due to the sub-pixel precision of its keypoint localization. However, when the threshold increased beyond 5 pixels, the accuracy of all methods, except SIFT and KAZE, exceeded 95\%, while SIFT's accuracy remains nearly constant. This suggests the limitations of SIFT's descriptor performance. When using AdaLAM for matching, as shown in Figure \ref{fig7}(b), the accuracy of SIFT, KAZE, ALIKED and DeDoDe decrease, while the accuracy of SuperPoint, R2D2, and DISK improve. By comparing this with the keypoint distributions in Figure \ref{fig5}, it can be observed that if the keypoints are evenly distributed, using AdaLAM can enhance accuracy. Conversely, if the keypoints are clustered in certain areas while absent in others, AdaLAM is likely to reduce accuracy, as the uneven distribution makes it difficult for AdaLAM to establish an affine model within local regions.

Regarding deep learning-based matching methods, as depicted in Figure \ref{fig7}(c), SuperGlue performs poorly, with its indoor model showing the lowest accuracy among all methods. LightGlue's performance is suboptimal when paired with SuperPoint, but results are similar when paired with SIFT, DISK, and ALIKED. Specifically, at a 4-pixel threshold, accuracy reaches 90\%, and approaches 100\% when the threshold exceeds 8 pixels. However, accuracy decreases for R2D2 and DISK. The three weights of LoFTR produce notably different results; accuracy approaches 100\% for thresholds above 5 pixels, with its outdoor model showing the highest accuracy below 5 pixels and its indoor model the lowest. Overall, DISK combined with AdaLAM and the LoFTR outdoor model perform best in this scenario, indicating their effectiveness in handling illumination changes and various disturbances.

\begin{figure*}
	\centering
	\includegraphics[width=2.0\columnwidth]{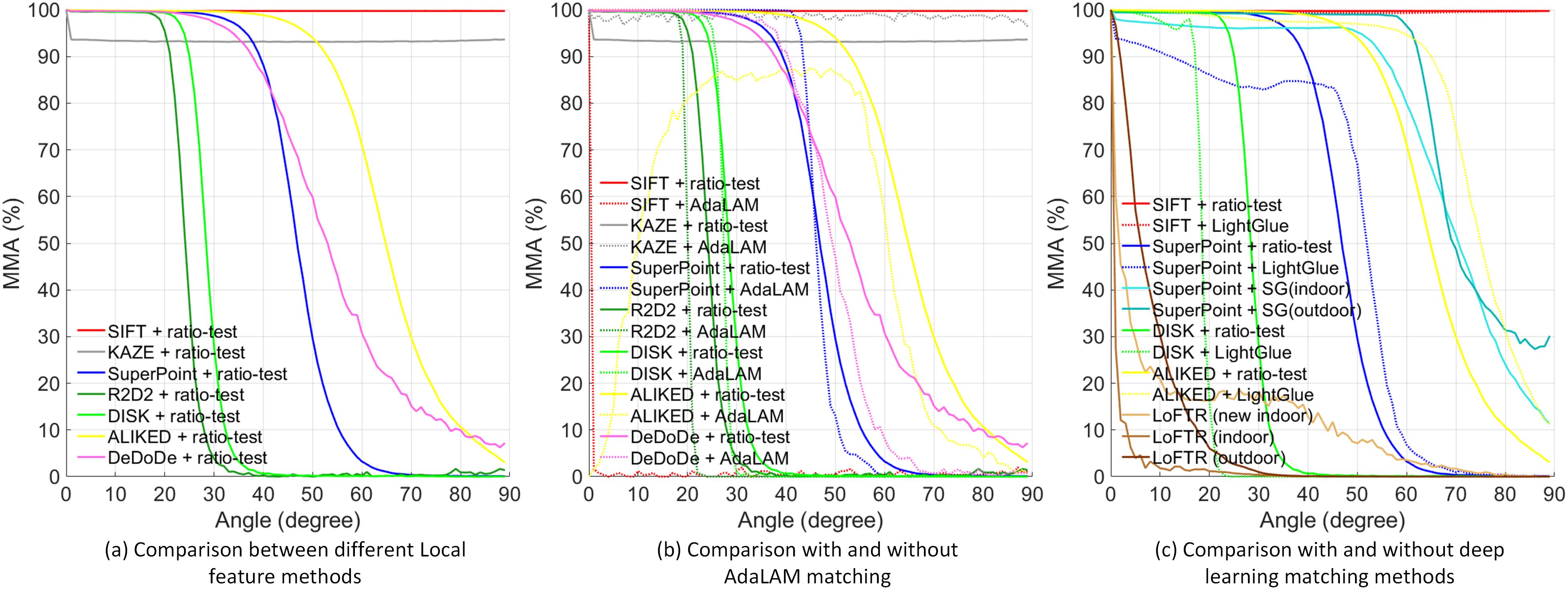}
	\caption{Mean Matching Accuracy (MMA) under various rotation angles.}
	\label{fig8}
\end{figure*}

Similarly, to evaluate the performance of various methods under different rotations, we match images with their copies rotated by various angles ranging from 0 to 90 degrees and calculate their MMA. The results are illustrated in Figure \ref{fig8}. When using the ratio-test for matching, as shown in Figure \ref{fig8}(a), SIFT achieves nearly 100\% accuracy consistently, indicating its excellent rotational invariance. In contrast, KAZE maintains an accuracy of around 93\% for most angles. Among deep learning-based feature methods, ALIKED significantly outperformed others, benefiting from its deformable convolutions. It achieves over 90\% accuracy up to a 50-degree rotation, whereas the accuracy of other methods falls below 40\%. The performance of R2D2 is worst, which has less than 5\% accuracy after a 30-degree rotation. Overall, the rotational invariance of deep learning methods remains relatively limited due to the inherent characteristics of vanilla convolution operations.

The analysis of matching results using AdaLAM, shown in Figure \ref{fig8}(b), reveals varying performance across different feature extraction methods. For SIFT, accuracy is near zero with almost no correct matches, likely due to the similar textures in coral reef images that hinder SIFT's ability to describe local regions accurately, causing AdaLAM to filter out matches. This indicates SIFT's limitations in complex, texture-rich underwater environments. In contrast, ALIKED exhibits low accuracy for small rotation angles but shows a significant improvement as the angle increases to around 30 degrees. This suggests that AdaLAM’s underlying assumptions may impose constraints on ALIKED’s matching ability at smaller rotations but become more suitable at larger angles. Among the deep learning-based methods, SuperGlue performs exceptionally well, maintaining over 95\% accuracy even at high rotation angles up to 50 degrees, demonstrating its robustness against rotation. SuperPoint and DISK combined with LightGlue show some reduction in accuracy at smaller angles, while ALIKED's accuracy improves with increased rotation, achieving over 70\% at angles exceeding 70 degrees. Overall, these results suggest that while traditional methods like SIFT struggle with the complex texture of coral imagery, deep learning methods such as SuperGlue provide more consistent performance across a range of conditions. ALIKED, in particular, shows promise in scenarios involving significant rotation, making it a potentially suitable choice for dynamic underwater environments.

\subsubsection{SfM reconstruction results evaluation}\label{E2S2}

Based on the correspondences obtained from image matching, SfM reconstruction can be performed. This study utilizes COLMAP (\cite{schonberger2016structure20}) as the SfM framework, inputting the matching results from Section \ref{E2S1} into COLMAP for reconstruction. The number of features extracted from each image is still restricted to 8000. For LoFTR, the results from its outdoor model are input into DF-SfM (\cite{he2024detector66}) for reconstruction. Additionally, the recently proposed end-to-end trained VGG-SfM (\cite{chen2019learning99}) is also evaluated. VGG-SfM supports the use of SIFT, SuperPoint, and ALIKED for extracting query points. However, due to its high memory consumption, VGG-SfM is not feasible for large-scale scenes like Coral-2018 and Coral-2019. In SfM reconstruction, a simple camera model with one focal length and two radial distortion parameters is applied to model the distortion. DF-SfM and VGG-SfM are constrained by GPU memory and therefore resize the longer edge of each image to 1200 pixels prior to processing. For other methods, we use the original images without additional processing is applied, though different methods may apply their own image processing. In this study, we follow the default settings provided by each original implementation.

Upon completion of SfM reconstruction, five evaluation metrics are computed: \textit{Rate}, \textit{Features}, \textit{Points}, \textit{Track}, and $E_{rep}$. \textit{Rate} represents the ratio of the number of images successfully registered to the total number of images. \textit{Features} denotes the average number of successfully matched features on aligned images. \textit{Points} refers to the number of 3D scene points generated by the SfM reconstruction (1k = 1000). \textit{Track} indicates the average number of 2D features corresponding to a single 3D scene point, which can be understood as the average number of observations per point. $E_{rep}$ is the average reprojection error of matched features.

\begin{figure}
	\centering
	\includegraphics[width=1.0\columnwidth]{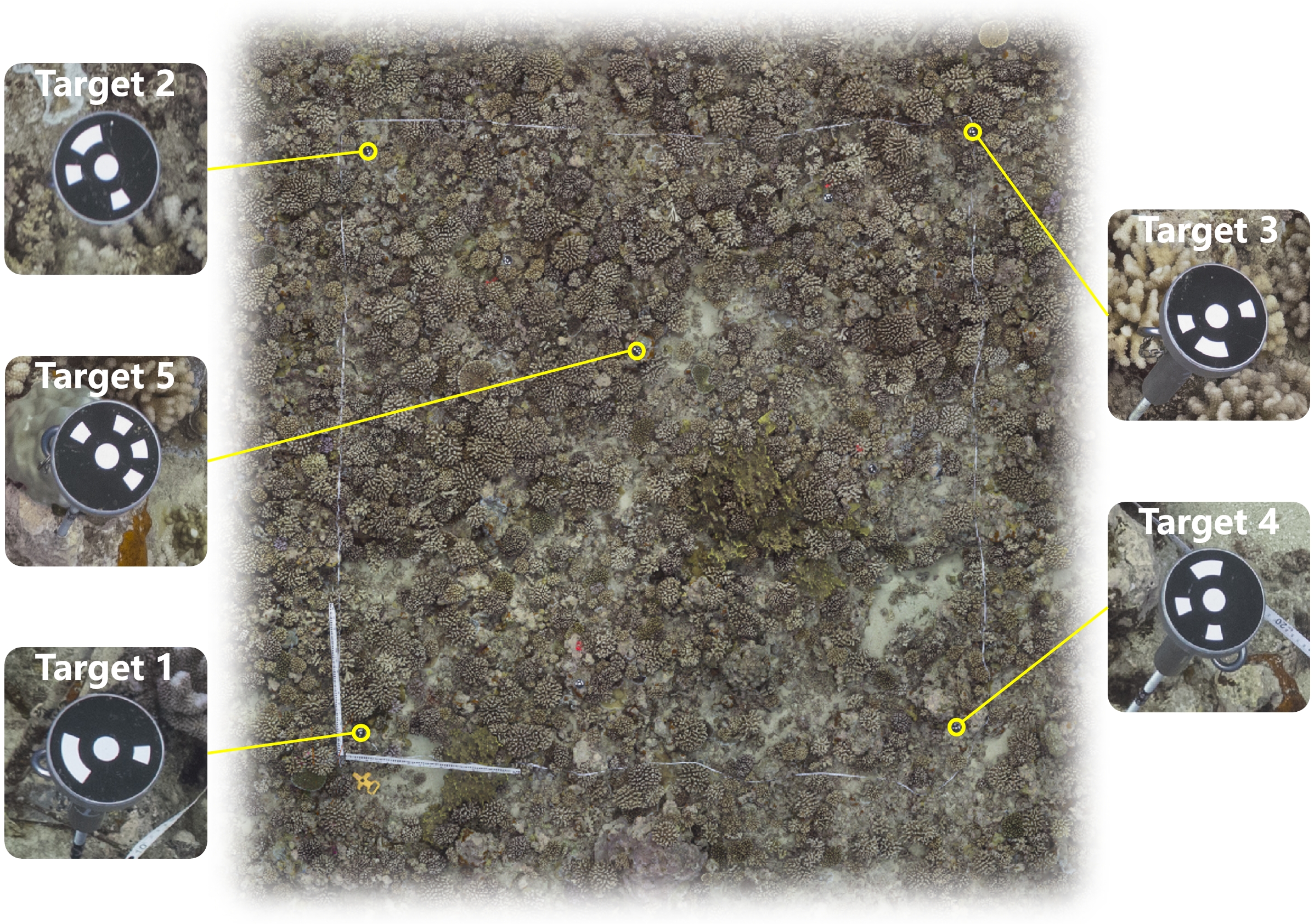}
	\caption{The distribution of GCPs in the Moorea Island survey area.}
	\label{fig89}
\end{figure}

For real-world datasets collected at Moorea Island (Coral-2018 and Coral-2019), five ground control points (GCPs) with planar photogrammetric coded targets were established, as illustrated in Figure \ref{fig89}. These targets, labelled Target 1 to Target 5, form the basis of an underwater geodetic control network. Their coordinates were determined using high-precision geodetic techniques (\cite{nocerino2020coral138}). Vertical heights were measured using geometric leveling with a green laser, while horizontal distances between points were carefully measured using a millimeter-graduated metal tape to support trilateration. The collected measurements were then processed through a network adjustment procedure to compute the final 3D coordinates of the GCPs. In this experiment, Target 5 is designated as a check point, and the remaining four as control points. Following initial SfM reconstruction, the GCP markers are identified across multiple images, enabling triangulation to estimate their positions in the arbitrary SfM coordinate system. To align the reconstructed model with the local geodetic control network, a seven-parameter transformation is estimated using the control points. Specifically, let $\mathbf{P}^{\mathrm{SfM}}_{\mathrm{ctrl},i}$ and $\mathbf{P}^{\mathrm{Local}}_{\mathrm{ctrl},i}$ denote the coordinates of the $i$-th control point in the SfM and local geodetic systems, respectively. The optimal transformation parameters, consisting of a scale factor $\hat{s}\in\mathbb{R}$, a rotation matrix $\hat{\mathbf{R}}\in\mathbb{R}^{3\times 3}$, and a translation vector $\hat{\mathbf{T}}\in\mathbb{R}^{3\times 1}$, are determined by solving the following least-squares minimization problem:
\begin{equation}
	(\hat{s}, \hat{\mathbf{R}}, \hat{\mathbf{T}}) 
	= \arg\min_{s, \mathbf{R}, \mathbf{T}} 
	\sum_{i \in C} \left\| s\,\mathbf{R}\,\mathbf{P}^{\mathrm{SfM}}_{\mathrm{ctrl},i} + \mathbf{T} - \mathbf{P}^{\mathrm{Local}}_{\mathrm{ctrl},i} \right\|^2_2
	\label{eq:seven-param}
\end{equation}
where $C$ is the set of control point indices. The spatial accuracy in object space is then evaluated by computing RMSEs between the transformed SfM coordinates and the ground-truth positions. Here, $\left\| \cdot \right\|_2$ denotes the 3D Euclidean norm, i.e., the square root of the sum of squared coordinate differences. RMSEs are reported separately for the control points $E_{\mathrm{control}}$ and the check point $E_{\mathrm{check}}$, both in millimeters. For each point $j$, the point-wise error is defined as the Euclidean distance between its transformed SfM coordinate $\mathbf{P}^{\mathrm{SfM}}_{j}$ and its ground-truth coordinate $\mathbf{P}^{\mathrm{Local}}_{j}$:
\begin{equation}
	E_j = \left\| \hat{s}\,\hat{\mathbf{R}}\,\mathbf{P}^{\mathrm{SfM}}_{j} + \hat{\mathbf{T}} - \mathbf{P}^{\mathrm{Local}}_{j} \right\|_2
	\label{eq:point-error}
\end{equation}
For synthetic datasets, the reconstructed camera positions are aligned to the ground truth positions from the simulation environment using a seven-parameter transformation. From this alignment, we compute the root mean square errors (RMSE) for camera positions $E_{loc}$ and orientations $E_{dir}$. The orientation errors are calculated based on the angular deviations between the reconstructed and ground-truth camera directions, measured in degrees. The quantitative metrics for the real-world datasets are presented in Table \ref{tab3}. For the simulation datasets, Coral UE5 and Coral UE5 (lite) are taken as examples, as shown in Table \ref{tab4}.

\begin{table*}[ht]
	\caption{Various metrics of the reconstruction results of Coral-2018 and Coral-2019 using different methods. The missing values, indicated by dashes, are due either to reconstruction failure or to the inapplicability of certain metrics to the corresponding methods.}
	\centering
	\label{tab3} 
	\renewcommand\arraystretch{1.1}
	\tabcolsep=0.1cm 
	\begin{adjustbox}{scale={0.9}, center} 
		\begin{tabular}{llllllllllllllll}
			\hline
			& & \multicolumn{7}{l}{\textbf{Coral-2018}} & \multicolumn{7}{l}{\textbf{Coral-2019}} \\
			Feature & Match & \textit{Rate} & \textit{Features} & \textit{Points} & \textit{Track} & ${E_{rep}}$ & ${E_{control}}$ & ${E_{check}}$ & \textit{Rate} & \textit{Features} & \textit{Points} & \textit{Track} & ${E_{rep}}$ & ${E_{control}}$ & ${E_{check}}$ \\
			\hline
			\multirow{3}{*}{SIFT} & RT & \textbf{451/451} & 4244.8 & 503,475 & 3.80 & 0.56 & 7.99 & 9.74 & \textbf{318/318} & 3946.6 & 324,741 & 3.86 & 0.51 & \textbf{3.35} & 41.37 \\
			& AdaLAM & \textbf{451/451} & 3799.5 & 338,350 & 5.06 & 0.72 & 8.05 & 10.71 & \textbf{318/318} & 3633.4 & 230,345 & 5.02 & 0.65 & 3.42 & 41.65 \\
			& LightGlue & \textbf{451/451} & 4457.1 & 338,517 & \textbf{5.94} & 1.07 & 7.73 & 11.84 & \textbf{318/318} & 4358.4 & 241,915 & \textbf{5.73} & 1.04 & 3.51 & 43.16 \\
			\arrayrulecolor{gray!30}\hline
			\multirow{2}{*}{KAZE} & RT & 447/451 & 3534.1 & 454,122 & 3.48 & \textbf{0.38} & 8.94 & 8.71 & 285/318 & 3324.0 & 302,276 & 3.13 & \textbf{0.33} & - & - \\
			& AdaLAM & \textbf{451/451} & 2924.1 & 272,200 & 4.84 & 0.51 & 9.08 & 10.54 & 314/318 & 2508.1 & 189,834 & 4.15 & 0.44 & 4.16 & 36.15 \\
			\hline
			\multirow{5}{*}{SuperPoint} & RT & 404/451 & 3426.8 & 400,455 & 3.46 & 0.65 & 57.39 & 70.00 & 312/318 & 3226.1 & 277,544 & 3.63 & 0.64 & 4.13 & 34.23 \\
			& AdaLAM & \textbf{451/451} & 3173.0 & 344,664 & 4.15 & 0.80 & 25.58 & 25.64 & 316/318 & 3140.9 & 233,431 & 4.25 & 0.77 & 4.40 & 36.81 \\
			& SG (indoor) & 229/451 & 4127.5 & 244,283 & 3.87 & 0.83 & - & - & 300/318 & 3839.5 & 290,361 & 3.97 & 0.84 & - & - \\
			& SG (outdoor) & \textbf{451/451} & 4777.7 & 527,298 & 4.09 & 0.97 & 10.72 & 5.54 & \textbf{318/318} & 4569.2 & 354,975 & 4.09 & 0.94 & 3.75 & 36.49 \\
			& LightGlue & \textbf{451/451} & 3654.5 & 529,035 & 3.12 & 1.35 & 8.48 & 5.41 & \textbf{318/318} & 3006.2 & 306,996 & 3.11 & 1.35 & 4.98 & 40.90 \\
			\hline
			\multirow{2}{*}{R2D2} & RT & 100/451 & 1724.1 & 38,071 & 4.53 & 1.09 & - & - & 190/318 & 1090.9 & 51,964 & 3.99 & 1.01 & 19.30 & 31.98 \\
			& AdaLAM & 122/451 & 2530.8 & 61,246 & 5.04 & 1.23 & 16.81 & 16.90 & 283/318 & 2211.0 & 142,993 & 4.38 & 1.16 & 16.92 & 45.54 \\
			\hline
			\multirow{3}{*}{DISK} & RT & 448/451 & 3826.8 & 394,511 & 4.35 & 1.02 & \textbf{7.39} & 3.81 & 307/318 & 3877.8 & 269,344 & 4.42 & 1.01 & 7.05 & 33.23 \\
			& AdaLAM & 448/451 & 4369.5 & 454,828 & 4.30 & 1.07 & 8.03 & \textbf{2.26} & 310/318 & 4330.9 & 306,455 & 4.38 & 1.06 & 5.04 & 34.29 \\
			& LightGlue & 46/451 & \textbf{5605.0} & 50,932 & 5.06 & 1.39 & - & - & 156/318 & 4556.0 & 177,453 & 4.01 & 1.32 & - & - \\
			\hline
			\multirow{3}{*}{ALIKED} & RT & \textbf{451/451} & 2968.4 & 319,944 & 4.18 & 0.50 & 8.45 & 4.13 & 317/318 & 2804.3 & 213,382 & 4.17 & 0.45 & 3.94 & 40.57 \\
			& AdaLAM & \textbf{451/451} & 3363.8 & 365,819 & 4.15 & 0.58 & 9.99 & 4.12 & 317/318 & 3229.7 & 247,330 & 4.14 & 0.54 & 3.63 & 40.59 \\
			& LightGlue & \textbf{451/451} & 4231.7 & 498,845 & 3.83 & 1.25 & 11.02 & 6.13 & \textbf{318/318} & 4665.9 & 370,183 & 4.01 & 1.27 & 4.23 & 37.91 \\
			\hline
			\multirow{2}{*}{DeDoDe} & RT & 447/451 & 2465.0 & 260,041 & 4.24 & 1.18 & 30.97 & 54.90 & 309/318 & 3120.2 & 221,701 & 4.35 & 1.23 & 5.07 & 38.64 \\
			& AdaLAM & \textbf{451/451} & 4654.1 & 426,499 & 4.92 & 1.36 & 42.32 & 22.33 & 317/318 & 4519.3 & 297,451 & 4.82 & 1.34 & 4.15 & 41.65 \\
			\hline
			\multicolumn{2}{l}{DF-SfM (LoFTR)} & \textbf{451/451} & 5193.0 & \textbf{1,019,030} & 2.30 & 0.45 & 37.92 & 52.55 & 312/318 & \textbf{5205.9} & \textbf{711,883} & 2.28 & 0.43 & 7.65 & \textbf{30.17} \\
			\arrayrulecolor{gray!100}\hline 
		\end{tabular}
	\end{adjustbox}
\end{table*}

\begin{table*}[ht]
	\caption{Various metrics of the reconstruction results of Coral-UE5 and Coral-UE5 (lite) using different methods. The missing values, indicated by dashes, are due either to reconstruction failure or to the inapplicability of certain metrics to the corresponding methods.}
	\centering
	\label{tab4}
	\renewcommand\arraystretch{1.1}
	\tabcolsep=0.15cm
	\begin{adjustbox}{scale={0.9}, center}
	\begin{tabular}{llllllllllllllll}
		\hline
		&        & \multicolumn{7}{l}{\textbf{Coral-UE5}}                  & \multicolumn{7}{l}{\textbf{Coral-UE5 (lite)}}                \\
		Feature               & Match  & \textit{Rate} & \textit{Features} & \textit{Points} & \textit{Track} & ${E_{rep}}$ & ${E_{loc}}$ & ${E_{dir}}$ & \textit{Rate} & \textit{Features} & \textit{Points} & \textit{Track} & ${E_{rep}}$ & ${E_{loc}}$ & ${E_{dir}}$ \\
		\hline
		\multirow{3}{*}{SIFT} & RT     & \textbf{70/70} & 3911.7   & 53,085  & 5.16  & \textbf{0.41}   & 1.18   & 2.00   & \textbf{18/18} & 953.9    & 5,564  & 3.09  & \textbf{0.35}   & 3.64   & 14.47  \\
		& AdaLAM & \textbf{70/70} & 4225.3   & 54,083  & 5.47  & 0.44   & 1.18   & 2.00   & \textbf{18/18} & 962.3    & 5,515  & 3.14  & 0.37   & 3.07   & 8.62   \\
		& LightGlue    & \textbf{70/70} & 5608.3   & 52,996  & 7.41  & 1.00   & 1.18   & 2.00   & \textbf{18/18} & 3108.1   & 16,429 & 3.41  & 0.91   & 1.20   & 3.25   \\
		\arrayrulecolor{gray!30}\hline
		\multirow{2}{*}{KAZE} & RT     & \textbf{70/70} & 3392.3   & 47,450  & 5.00  & 0.57   & 4.11   & 13.53  & \textbf{18/18} & 556.0    & 3,536  & 2.83  & 0.50   & 1.20   & 3.23   \\
		& AdaLAM & \textbf{70/70} & 4196.0   & 52,518  & 5.59  & 0.66   & 1.18   & 2.00   & \textbf{18/18} & 634.1    & 3,640  & 3.14  & 0.61   & 1.22   & 3.93   \\
		\hline
		\multirow{5}{*}{SuperPoint} & RT     & \textbf{70/70} & 3292.3   & 43,856  & 5.25  & 0.81   & 1.18   & 2.00   & 0/18  & -        & -      & -     & -      & -      & -      \\
		& AdaLAM       & \textbf{70/70} & 4822.4   & 53,241  & 6.34  & 1.02   & 1.18   & 2.00   & 11/18 & 488.0    & 1,785  & 3.01  & 0.73   & 5.62   & 18.31  \\
		& SG (indoor)  & \textbf{70/70} & 6370.6   & 61,987  & 7.19  & 1.18   & 1.18   & 1.99   & \textbf{18/18} & 2366.4   & 11,856 & 3.59  & 1.06   & 1.20   & \textbf{3.22 }  \\
		& SG (outdoor) & \textbf{70/70} & {7030.5}   & 60,582  & 8.12  & 1.27   & 1.18   & 1.99   & \textbf{18/18} & \textbf{4232.0}   & \textbf{19,942} & 3.82  & 1.13   & 1.20   & 3.23   \\
		& LightGlue    & \textbf{70/70} & 2168.3   & 27,535  & 5.51  & 1.44   & 1.18   & 2.00   & \textbf{18/18} & 1151.7   & 6,370  & 3.25  & 1.28   & 1.20   & 3.23   \\
		\hline
		\multirow{2}{*}{R2D2}       & RT           & \textbf{70/70} & 2268.1   & 31,322  & 5.07  & 0.76   & 1.18   & \textbf{1.98 }  & 0/18  & -        & -      & -     & -      & -      & -      \\
		& AdaLAM       & \textbf{70/70} & 3938.5   & 38,021  & 7.25  & 0.90   & 1.18   & 1.98   & 0/18  & -        & -      & -     & -      & -      & -      \\
		\hline
		\multirow{3}{*}{DISK}       & RT           & \textbf{70/70} & 5236.6   & 54,810  & 6.69  & 0.93   & 1.18   & 1.99   & 0/18  & -        & -      & -     & -      & -      & -      \\
		& AdaLAM       & \textbf{70/70} & 6179.2   & 61,476  & 7.04  & 1.05   & 1.18   & 1.99   & \textbf{18/18} & 1231.6   & 7,062  & 3.14  & 0.87   & 7.19   & 57.58  \\
		& LightGlue    & 50/70 & 4705.3   & 45,323  & 5.19  & 1.23   & 1.18   & 2.11   & 0/18  & -        & -      & -     & -      & -      & -      \\
		\hline
		\multirow{3}{*}{ALIKED}     & RT           & \textbf{70/70} & 6022.1   & 54,217  & 7.78  & 0.61   & 1.18   & 2.00   & \textbf{18/18} & 1528.9   & 7,814  & 3.52  & 0.52   & \textbf{1.19}   & 3.31   \\
		& AdaLAM       & \textbf{70/70} & 6785.8   & 54,633  & 8.69  & 0.76   & 1.18   & 2.00   & \textbf{18/18} & 2374.4   & 11,574 & 3.69  & 0.65   & 1.19   & 3.26   \\
		& LightGlue    & \textbf{70/70} & 3373.4   & 26,766  & 8.82  & 1.21   & 1.18   & 2.00   & \textbf{18/18} & 2272.8   & 10,688 & 3.83  & 1.12   & 1.20   & 3.24   \\
		\hline
		\multirow{2}{*}{DeDoDe}     & RT           & \textbf{70/70} & 2035.2   & 26,776  & 5.32  & 0.71   & 1.18   & 1.99   & 0/18  & -        & -      & -     & -      & -      & -      \\
		& AdaLAM       & \textbf{70/70} & 6188.5   & 39,653  & 10.92 & 1.18   & 1.18   & 2.00   & 17/18 & 1539.2   & 6,915  & 3.78  & 0.94   & 5.27   & 18.27  \\
		\hline
		\multicolumn{2}{l}{DF-SfM (LoFTR)}         & \textbf{70/70} & 4926.7   & {114,733} & 3.01  & 0.52   & 1.18   & 1.99   & \textbf{18/18} & 2152.4   & 15,834 & 2.45  & 0.43   & 3.66   & 22.20  \\
		\hline
		\multicolumn{2}{l}{VGG-SfM (SIFT)}         & \textbf{70/70} & 1982.0   & 2,892   & {47.97} & -      & {1.18}   & 2.00   & \textbf{18/18} & 3049.5   & 4,539  & \textbf{12.09} & -      & 1.21   & 3.27   \\
		\multicolumn{2}{l}{VGG-SfM   (SuperPoint)} & \textbf{70/70} & 1959.2   & 3,983   & 34.43 & -      & 1.18   & 2.00   & \textbf{18/18} & 1477.3   & 3,148  & 8.45  & -      & 1.24   & 3.28   \\
		\multicolumn{2}{l}{VGG-SfM (ALIKED)}       & \textbf{70/70} & 2780.1   & 4,977   & 39.10 & -      & 1.18   & 2.01   & \textbf{18/18} & 2889.7   & 5,445  & 9.55  & -      & 1.20   & 3.24  \\
		\arrayrulecolor{gray!100}\hline
	\end{tabular}
\end{adjustbox}
\end{table*}

In general, R2D2 exhibits the weakest performance, particularly on the real-world datasets, where many images fail to be successfully aligned and the resulting errors remain at the centimeter level, falling short of the sub-centimetric accuracy for precise modeling. It also fails to achieve any meaningful reconstruction on the Coral-UE5 (lite) dataset. In comparative analyses, R2D2 consistently yields the lowest values for \textit{Features} and \textit{Points}. This suboptimal performance is likely due to the R2D2 features' significant degradation under moderate changes in viewpoint. In contrast, ALIKED exhibits significantly better performance. It achieves the highest overall \textit{Rate} for reconstruction and demonstrates lower values across various error metrics, indicating that its results are not only complete and reliable but also accurate. SIFT generally produces favorable results, although its performance is suboptimal when handling Coral-UE5 (lite) with the ratio-test. For synthetic datasets, different approaches often produce similar $E_{loc}$ and $E_{dir}$ because most feature matching strategies can obtain sufficient correct correspondences under standard conditions. Furthermore, the SfM pipeline's outlier removal helps maintain reconstruction accuracy by eliminating significant outliers, and performance differences become apparent only when image overlap decreases. Approaches struggling to establish an adequate number of correct correspondences lead to either complete reconstruction failure or significantly increased errors, as evidenced by the performance degradation of multiple approaches on the Coral-UE5 (lite) dataset. For more complex real-world datasets, SIFT, DISK and ALIKED achieve millimeter-level accuracy on Coral-2018, as reflected in their low $E_{control}$ and $E_{check}$. SuperPoint achieves high accuracy only when paired with learning-based matching methods and otherwise suffers from large errors. DeDoDe produces large errors despite performing well on synthetic datasets, suggesting limited robustness for practical high-precision coral reef modeling. As for Coral-2019, the sparse coverage of the check point in the collected imagery results in generally larger errors. Nevertheless, the residuals at the control points reveal that most methods perform comparably, except for R2D2, whose errors are the largest and render it nearly unusable.

In terms of feature matching, AdaLAM often proves beneficial in enhancing the completeness of the reconstruction. It effectively filters out significant errors and identifies potential correspondences, thereby increasing the \textit{Track} value and contributing to a more stable SfM reconstruction. This is particularly relevant in real-world scenarios where the complexity of image acquisition often leads to challenging matching conditions and a higher likelihood of mismatches. Nevertheless, it is crucial to recognize that the use of AdaLAM does not necessarily improve accuracy and may, in fact, slightly increase the error in many cases. It can lead to an increase in reprojection error because the features matched by AdaLAM might not correspond precisely to the same 3D scene point but rather approximate locations. This can be considered as a trade-off between robustness and reliability versus precision.

As for SuperGlue, its outdoor model demonstrates excellent performance. Even though it might occasionally fail to match some image pairs, it is generally effective at aligning all images and achieves high \textit{Feature} values, indicating strong performance under the specified conditions. Conversely, its indoor model performs well on Coral-UE5, but underperforms on Coral-2018, with fewer than half of the images successfully aligned. While a limited number of mismatches can be filtered out during the reconstruction process, an excessive number of mismatches poses significant problems. In contrast, LightGlue performs exceptionally well when combined with SIFT or ALIKED, showcasing excellent results. However, its performance is less effective when paired with SuperPoint or DISK. This outcome is largely consistent with the previously obtained image matching results. The results of LoFTR-based DF-SfM are quite distinct, with a high number of Points and a low \textit{Track} value. This discrepancy can be attributed to its detector-free design, which establish correspondences between images without relying on explicit keypoints, thereby limiting the potential for repeated observations of the same 3D point. Moreover, its iterative refinement process contributes to the low reprojection errors observed. Although DF-SfM achieves relatively high precision on the Coral-UE5 dataset, it suffers from significant errors on Coral-UE5 (lite), indicating challenges in reconstruction and limited robustness under more constrained conditions. Furthermore, despite its very low reprojection errors on Coral-2018, it exhibits large ground control point (GCP) errors, suggesting inconsistencies in global alignment. The results for VGG-SfM are notably different from those of DF-SfM, with VGG-SfM exhibiting a low \textit{Points} value and a high \textit{Track} value. This is because that VGG-SfM utilizes deep 2D point tracking to establish correspondences between images. The performance of VGG-SfM on simulation datasets is comparable to that of standard SfM, indicating its viability. However, due to its end-to-end design, VGG-SfM incurs substantial memory overhead, which restricts its use for larger-scale reconstructions. In summary, while SIFT, as a widely-used method, demonstrates solid performance, the learning-based ALIKED generally surpass it, and they are both effective choices. AdaLAM and LightGlue show strong applicability and are also reliable options, though their effectiveness may vary depending on the chosen feature extraction method.

\subsection{Dense surface reconstruction results comparison}\label{E3}

Based on the accurate camera poses, dense reconstruction techniques can be employed to capture the intricate geometric structures of coral reefs. This section presents a comparative evaluation of the four categories of dense reconstruction methods discussed in Section \ref{dense}, focusing on their reconstruction fidelity, accuracy, and efficiency. We implement representative approaches from each category of methods. For the first category, traditional MVS methods, we use the dense reconstruction functionality of COLMAP (\cite{schonberger2016structure20}), which employs a depth map-based MVS algorithm (\cite{schonberger2016pixelwise46}). In the second category, deep learning-based MVS, we employ Vis-MVSNet (\cite{zhang2023vis27}) and MVSFormer++ (\cite{cao2024mvsformer++52}). For the third category, NeRF-based methods, we apply Instant-NGP (\cite{muller2022instant15}), Nerfacto (\cite{tancik2023nerfstudio22}), and Neuralangelo (\cite{li2023neuralangelo12}). Our preliminary experiments indicate that NeuS2 (\cite{wang2023neus2_85}) and BakedSDF (\cite{yariv2023bakedsdf87}) are unable to reconstruct coral reef scenes and are therefore excluded from further comparison. Finally, for the fourth category, GS-based methods, we apply SuGaR (\cite{guedon2024sugar30}), 2D GS (\cite{huang20242d32}), and GOF (\cite{yu2024gaussian86}).

It is important to note that different methods produce outputs in various formats (refer to Table \ref{tab2} in Section \ref{dense}). For Instant-NGP and Nerfacto, point clouds can be exported using Nerfstudio. The resulting point clouds can then be used with Screened Poisson Surface Reconstruction to generate mesh models (\cite{kazhdan2013screened88}). For MVSFormer++, there is a pre-trained model trained on DTU (\cite{aanaes2016large77}) as well as a model fine-tuned on Tanks-and-Temples (\cite{knapitsch2017tanks78}), the latter of which performs better overall in our pre-experiments and is therefore used for subsequent comparative analysis. While NeRF-based and GS-based methods have shown promise in dense surface reconstruction, their direct application to large-scale scenes, such as Coral-2018 and Coral-2019, can be challenging. Reconstruction of these scenes can instead be accomplished through MVS methods or by selecting images of a specific region. As for the experimental environment, all experiments in this section were conducted using an NVIDIA GeForce RTX 3090 GPU, and all image data are scaled to a maximum dimension of 1600 pixels. COLMAP is implemented in C++, while the other algorithms are implemented in Python. Specifically, Instant-NGP and Nerfacto were executed using the Nerfstudio framework.

\subsubsection{Reconstruction quality evaluation}\label{E3S1}

\begin{figure*}
	\centering
	\includegraphics[width=2.0\columnwidth]{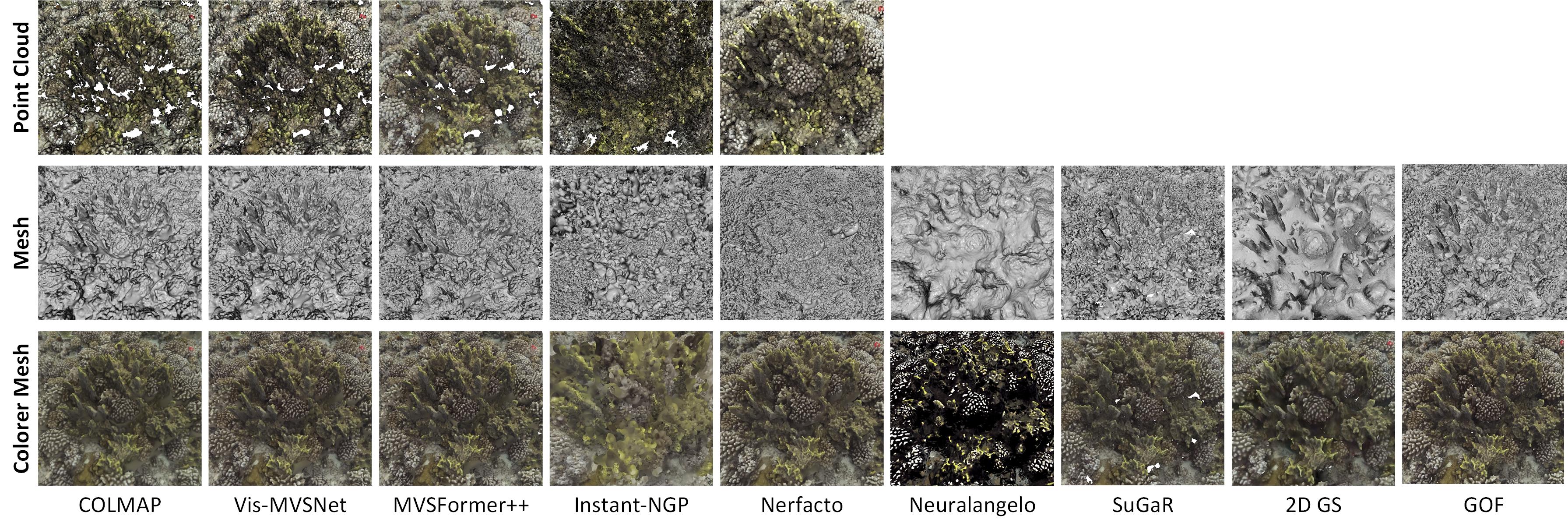}
	\caption{Comparison of dense reconstruction results for Coral-2018 (partial).}
	\label{fig9}
\end{figure*}

\begin{figure*}
	\centering
	\includegraphics[width=2.0\columnwidth]{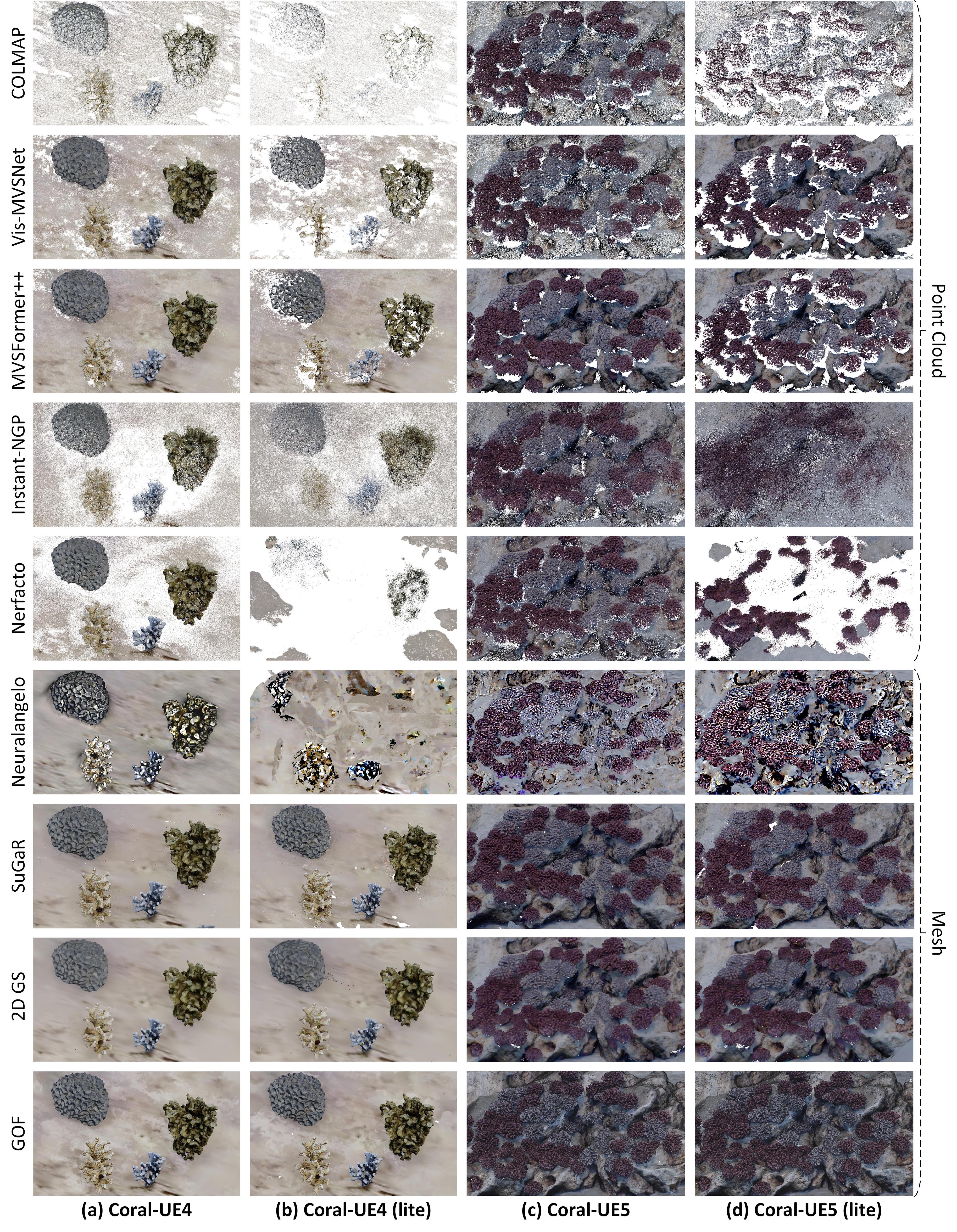}
	\caption{Comparison of dense reconstruction results for synthesis datasets. Because of the significant variability in meshes generated from dense point clouds depending on the methods used, we present the point clouds directly output by COLMAP, Vis-MVSNet, MVSFormer++, Instant-NGP, and Nerfacto, rather than the meshes.}
	\label{fig10}
\end{figure*}

For the evaluation of reconstruction effects and accuracy, images from a specific region of the Coral-2018 dataset (a total of 42 images) are used for real-world scenario experiments, referred to as Coral-2018 (partial). Simulated data, on the other hand, are used to quantitatively assess reconstruction accuracy and the impact of image acquisition density on reconstruction. The corresponding dense reconstruction results are presented in Figure \ref{fig9} and Figure \ref{fig10}, respectively.

Figure \ref{fig9} presents the point clouds, meshes, and colored meshes produced by each solution. It should be noted that the meshes generated by SuGaR are produced through texture mapping. In general, COLMAP, Vis-MVSNet, and MVSFormer++ yield comparable results. However, COLMAP shows limited detail recovery, and its mesh reconstruction is relatively smooth, failing to capture the intricate structures of coral tentacles. In contrast, MVSFormer++ provides the most detailed and accurate reconstructions, with a clear depiction of coral structures and minimal noise, despite its training dataset not including coral reef scenes. The results from Instant-NGP are overall the least effective, with the point clouds exhibiting a significant number of outliers, making surface reconstruction difficult. Although the results of Nerfacto exhibit some outliers, their quantity is considerably reduced, resulting in a model with a slightly uneven surface but fine geometric details, including sharp reef features. Conversely, the surface generated by Neuralangelo is overly smooth, failing to capture the intricate structures of coral reefs. The models reconstructed by SuGaR exhibit good visual quality; however, they fall short in accurately representing fine details. Specifically, it uses coarse meshes for smooth areas in the real world, which does not meet the high-precision measurement requirements. This limitation is partly due to its significant computational overhead, resulting in less dense meshes. The 2D GS method produces surfaces that are generally smooth, similar to those generated by Neuralangelo. However, 2D GS excels in areas with favorable lighting and rich textures, such as the protruding sections of rocks. In contrast, it struggles with accurate reconstruction in poorly lit regions, such as shadows within rocks and gaps between coral tentacles. Among these methods, GOF demonstrates the best performance. Its mesh is smooth, complete, and detailed, successfully representing the intricate structures of coral reefs.

The direct reconstruction results of various methods on the synthetic image datasets are presented in Figure \ref{fig10}, which includes Coral-UE4, Coral-UE5, and their lite versions. For the Coral-UE4 and Coral-UE5 datasets, the reconstruction results are comparable to those shown in Figure \ref{fig9}. MVSFormer++ produces the most dense point clouds, while Instant-NGP generates sparser point clouds with substantial noise. Both Neuralangelo and 2D GS generate relatively smooth surfaces, whereas GOF exhibits the most superior performance, yielding the most refined and accurate mesh models. In the Coral-UE4 (lite) and Coral-UE5 (lite) datasets, which contain only a quarter of the images compared to the original datasets, the reconstruction results from all solutions exhibit varying degrees of degradation, with noticeable differences in performance. Specifically, COLMAP produces sparser point clouds, with many areas missing points. Although the point clouds reconstructed by Vis-MVSNet and MVSFormer++ are also less complete than those obtained using the full datasets, they are still able to effectively recover the intricate geometric structures of the coral reefs to a reasonable extent. NeRF-based methods exhibit significant performance degradation under limited data conditions. Instant-NGP produces point clouds with a marked increase in noise, impairing the accurate representation of scene geometry. Nerfacto struggles to generate usable point clouds, and Neuralangelo encountered severe errors with the Coral-UE4 (lite) dataset, failing to reconstruct the coral reefs. These issues highlight that NeRF-based approaches depend heavily on datasets with high overlap. When data is sparse, their performance deteriorates sharply, likely due to insufficient data for effective model fitting. In contrast, GS-based methods demonstrate robust performance even with reduced data. SuGaR, 2D GS, and GOF successfully reconstructed the coral reef with high accuracy and completeness. Notably, GOF effectively captured fine coral structures, suggesting that GS-based methods are reliable alternatives when data is limited.

\begin{figure*}
	\centering
	\includegraphics[width=2.0\columnwidth]{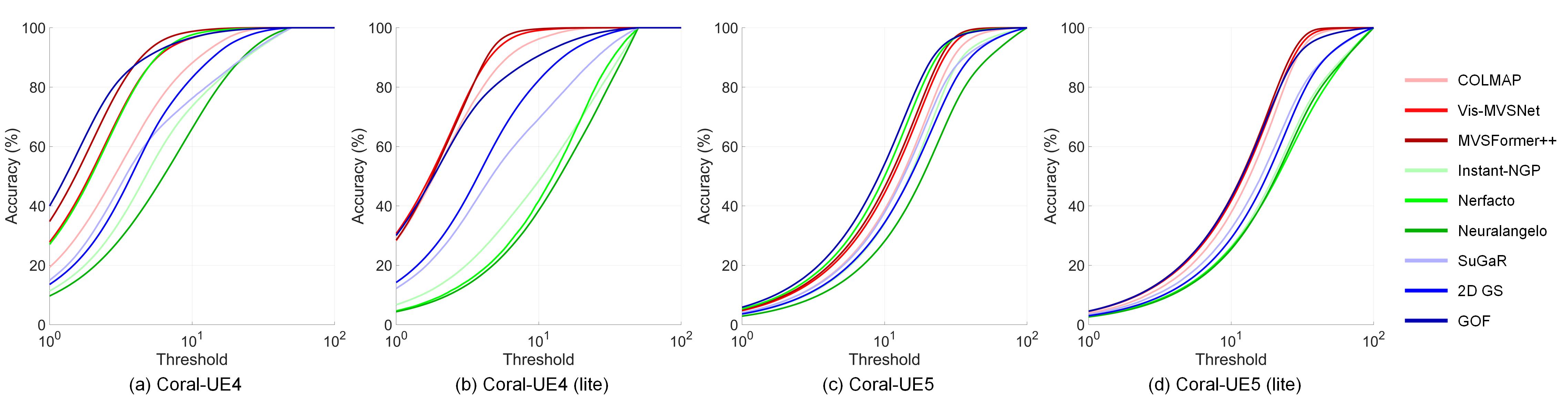}
	\caption{Accuracy of the reconstruction results for each method at different distance thresholds.}
	\label{fig11}
\end{figure*}

\begin{figure*}
\centering
\includegraphics[width=2.0\columnwidth]{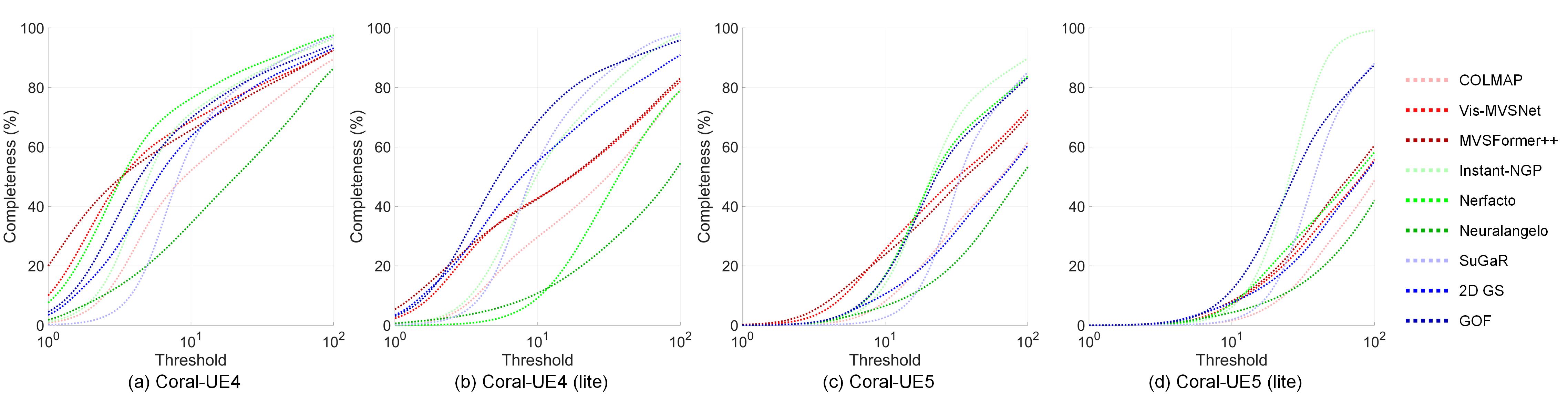}
\caption{Completeness of the reconstruction results for each method at different distance thresholds.}
\label{fig12}
\end{figure*}

\begin{figure*}
\centering
\includegraphics[width=2.0\columnwidth]{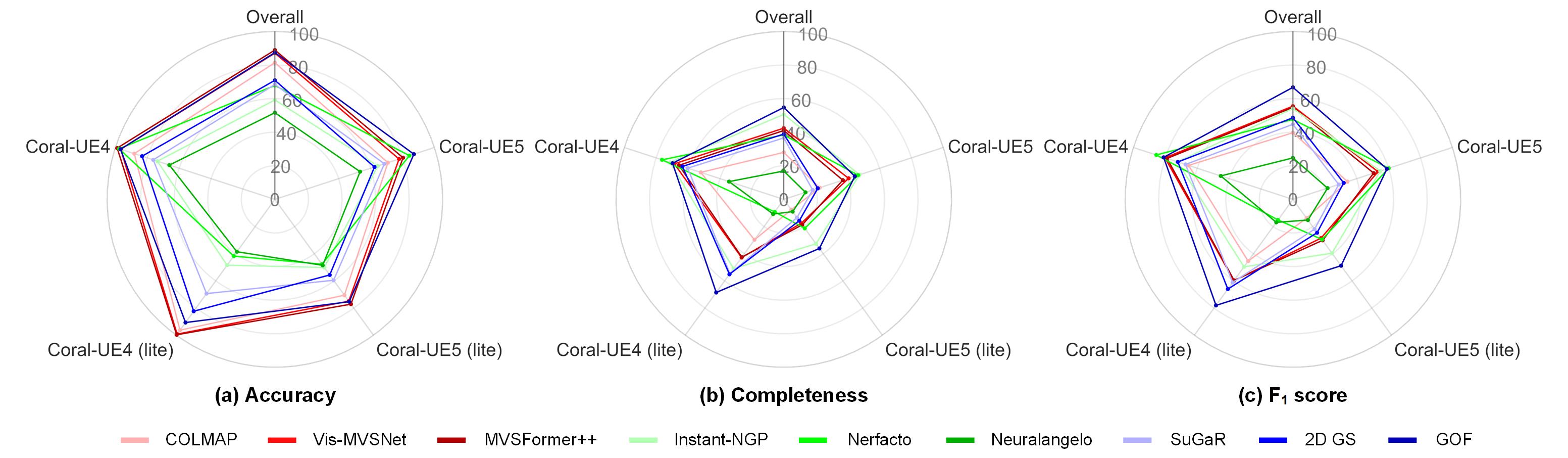}
\caption{Radar plots illustrating the results of various dense reconstruction methods across four different datasets in terms of (a) accuracy (\%), (b) completeness (\%), and (c) $F_1$ score (\%). "Overall" represents the average value for each metric.}
\label{fig13}
\end{figure*}

To quantitatively assess the reconstruction accuracy of various methods, we align the 3D points obtained from each method with the ground-truth model from the simulation environments using the Iterative Closest Point (ICP) algorithm. Following the previous work (\cite{schops2017multi131}), the reconstruction results are evaluated using two metrics: \textit{accuracy} and \textit{completeness}. Both measures are evaluated over a range of distance thresholds from 1 to 100. \textit{Accuracy} indicates how closely the reconstructed points align with the ground-true surface, while \textit{completeness} measures how much of the actual scene is captured. High \textit{accuracy} ensures that details are faithfully represented, and high \textit{completeness} ensures that model covers all important features — both are essential for a reliable 3D reconstruction. Specifically, in this study, \textit{accuracy} is defined as the percentage of reconstructed 3D points whose distance to the ground-truth mesh is below a given threshold. This distance is calculated using CloudCompare (v2.13.2) with the "Cloud-to-Mesh Distance" tool, which measures the distance from each point in the point cloud to the nearest triangle on the reference mesh. \textit{Completeness}, on the other hand, is determined by measuring the distance from each ground-truth point to the nearest reconstructed point and is defined as the proportion of ground-truth points that fall within the specified distance threshold. Moreover, given the importance of both metrics, the $F_1$ score can serve as a comprehensive single measure to rank the results. The $F_1$ score is defined as the harmonic mean of \textit{accuracy} (precision) $p$ and \textit{completeness} (recall) $r$, calculated as $2 \cdot (p \cdot r)/(p+r)$. The results of each method under different thresholds are shown in Figure \ref{fig11} and Figure \ref{fig12}. Additionally, Figure \ref{fig13} displays radar plots of the accuracy, completeness, and $F_1$ score for each method. These plots are based on a distance threshold of 10 for the Coral-UE4 and Coral-UE4 (lite) datasets, and a threshold of 20 for the Coral-UE5 and Coral-UE5 (lite) datasets. Additionally, the comparative performance of various methodologies is visualized in Figure 14 through radar plots displaying three key evaluation metrics: accuracy, completeness, and $F_1$ score. To accommodate the variability in camera-to-object distances across experimental scenarios, different distance thresholds are established: 10 for the Coral-UE4 and Coral-UE4 (lite) datasets, and 20 for the Coral-UE5 and Coral-UE5 (lite) datasets.

Overall, GOF proves to be the most reliable method for coral reef reconstruction, generally outperforming other approaches. It delivers both high accuracy and completeness, even when data is limited. Vis-MVSNet and MVSFormer++ perform slightly lower than GOF but still achieve high accuracy and good completeness, particularly achieving superior precision in scenarios with fewer images. NeRF-based methods exhibit adequate performance when data is abundant, with Nerfacto achieving the highest $F_1$ scores on both Coral-UE4 and Coral-UE5. However, when the number of images decreases, these methods suffer from substantial performance degradation, with accuracy falling behind both MVS-based and GS-based methods, indicating their struggle to produce reliable results under data-constrained conditions. Furthermore, it is noteworthy that while the relative accuracy rankings among methods remain largely consistent across varying thresholds, certain methods, particularly Instant-NGP and SuGaR, exhibit significant increases in completeness as the threshold increases. This pattern suggests that these methods generate numerous points in proximity to the ground truth, but with lower precision. In contrast, methods such as MVSFormer++, which demonstrate slower growth in completeness scores, achieve more accurate reconstruction but may fail to capture certain structural elements of the scene, potentially resulting in reconstruction gaps.

\subsubsection{Efficiency}\label{E3S2}

Dense surface reconstruction is typically one of the most time-consuming steps in 3D reconstruction. We calculate the processing time for each method across different datasets, as shown in Table \ref{tab5}. Since methods other than COLMAP, Vis-MVSNet, and MVSFormer++ are not yet suitable for reconstructing large-scale scenes, their processing times for Coral-2018 and Coral-2019 were not included in the comparison. It can be observed that Vis-MVSNet and MVSFormer++ are significantly faster than the other methods, taking only a few minutes to process datasets containing dozens of images. In contrast, NeRF-based and GS-based methods require much longer time due to their optimization process, with runtime dependent on the number of training iterations. In this preliminary comparison, the number of iterations for each method was fixed. Although more iterations generally result in better model fitting, the benefits diminish over time, making it necessary to adjust training duration according to specific needs. The efficiency comparison in this paper is preliminary and may change as research progresses. Traditional MVS- and NeRF-based methods are continuously being improved for efficiency, and future research will also focus on enhancing the efficiency of GS-based methods.

\begin{table*}[]
	\caption{The runtime in seconds of different methods (in seconds).}
	\centering
	\label{tab5}
	\renewcommand\arraystretch{1.1}
\begin{adjustbox}{scale={0.9}, center}
	\begin{tabular}{llllllll}
		\hline
		Methods      & Coral-2018 & Coral-2019 & Coral-2018 (partial) & Coral-UE4 & Coral-UE4 (lite) & Coral-UE5 & Coral-UE5 (lite) \\
		\hline
		COLMAP       & 13565      & 9531       & 1287                 & 1910      & 457              & 1858      & 463              \\
		Vis-MVSNet   & 1817       & 1102       & \textbf{155}         & 183       & 77               & 208       & 81               \\
		MVSFormer++  & \textbf{1311}  & \textbf{697}   & 164          & \textbf{172}  & \textbf{59}  & \textbf{199}  & \textbf{61}               \\
		Instant-NGP  & -          & -          & 1309                 & 1208      & 1253             & 1254      & 1210             \\
		Nerfacto     & -          & -          & 1225                 & 1145      & 1322             & 1054      & 1128             \\
		Neuralangelo & -          & -          & 20372                & 20251     & 20544            & 20689     & 20003            \\
		SuGaR        & -          & -          & 9044                 & 6425      & 5059             & 5373      & 5184             \\
		2D GS        & -          & -          & 14062                & 6764      & 6017             & 8048      & 7480             \\
		GOF          & -          & -          & 13790                & 5889      & 6193             & 7422      & 7589            \\
		\hline
	\end{tabular}
\end{adjustbox}
\end{table*}

\begin{figure}
	\centering
	\includegraphics[width=0.85\columnwidth]{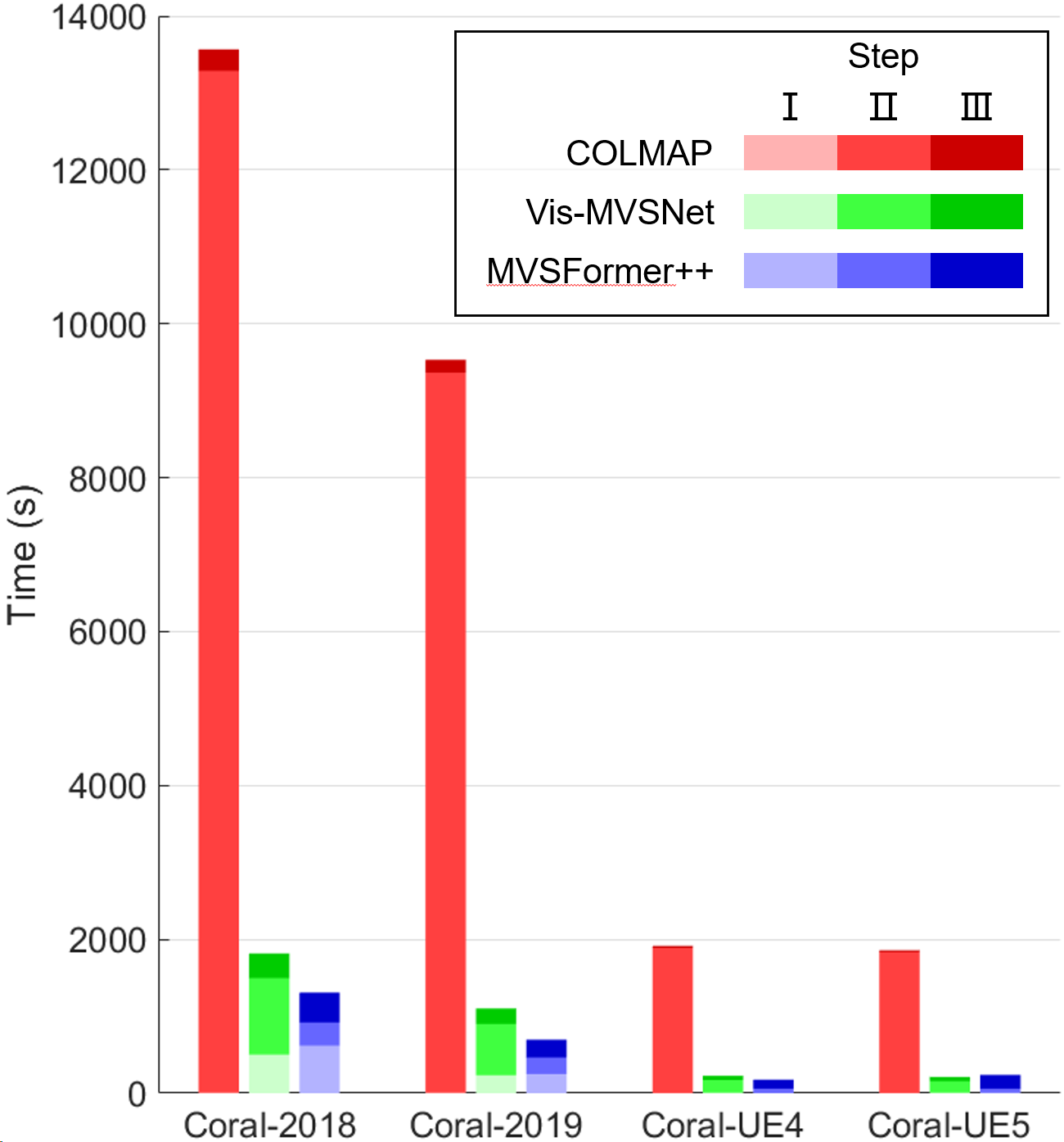}
	\caption{The runtime of each step in the MVS methods.}
	\label{fig14}
\end{figure}

Additionally, to further investigate the efficiency of MVS methods when reconstructing large-scale scenes, we analyze the processing steps of COLMAP, Vis-MVSNet, and MVSFormer++. These three methods, all based on depth map generation in MVS, generally involve three main steps: I. Data Preparation: This includes operations like data format conversion; II. Depth Map Generation: Depth maps are generated using multi-view images; III. Fusion and Filtering: Depth maps are fused to create the dense point cloud, followed by filtering to remove outliers. Figure \ref{fig14} illustrates the runtime of these three methods across multiple datasets. It is evident that the runtime of COLMAP is predominantly consumed by the depth map generation step, which takes several times longer than that of deep learning-based MVS methods. This extended runtime is due to COLMAP's iterative process for refining disparity maps. In contrast, Vis-MVSNet and MVSFormer++, as end-to-end deep learning networks, generate depth maps through feed-forward propagation, which is significantly faster. However, these methods exhibit longer runtimes during the data pre-processing and final fusion stages, indicating potential areas for future optimization.

In summary, considering the reconstruction quality and accuracy results presented in Section \ref{E3S1}, MVSFormer++ is the most practical method for reconstructing large-scale scenes, offering both high precision and rapid processing. For smaller-scale scenes, methods based on NeRF or GS, such as GOF, may present a viable alternative.

\subsection{Comparison with commercial 3D reconstruction software}\label{software}

\begin{figure*}
	\centering
	\includegraphics[width=2.0\columnwidth]{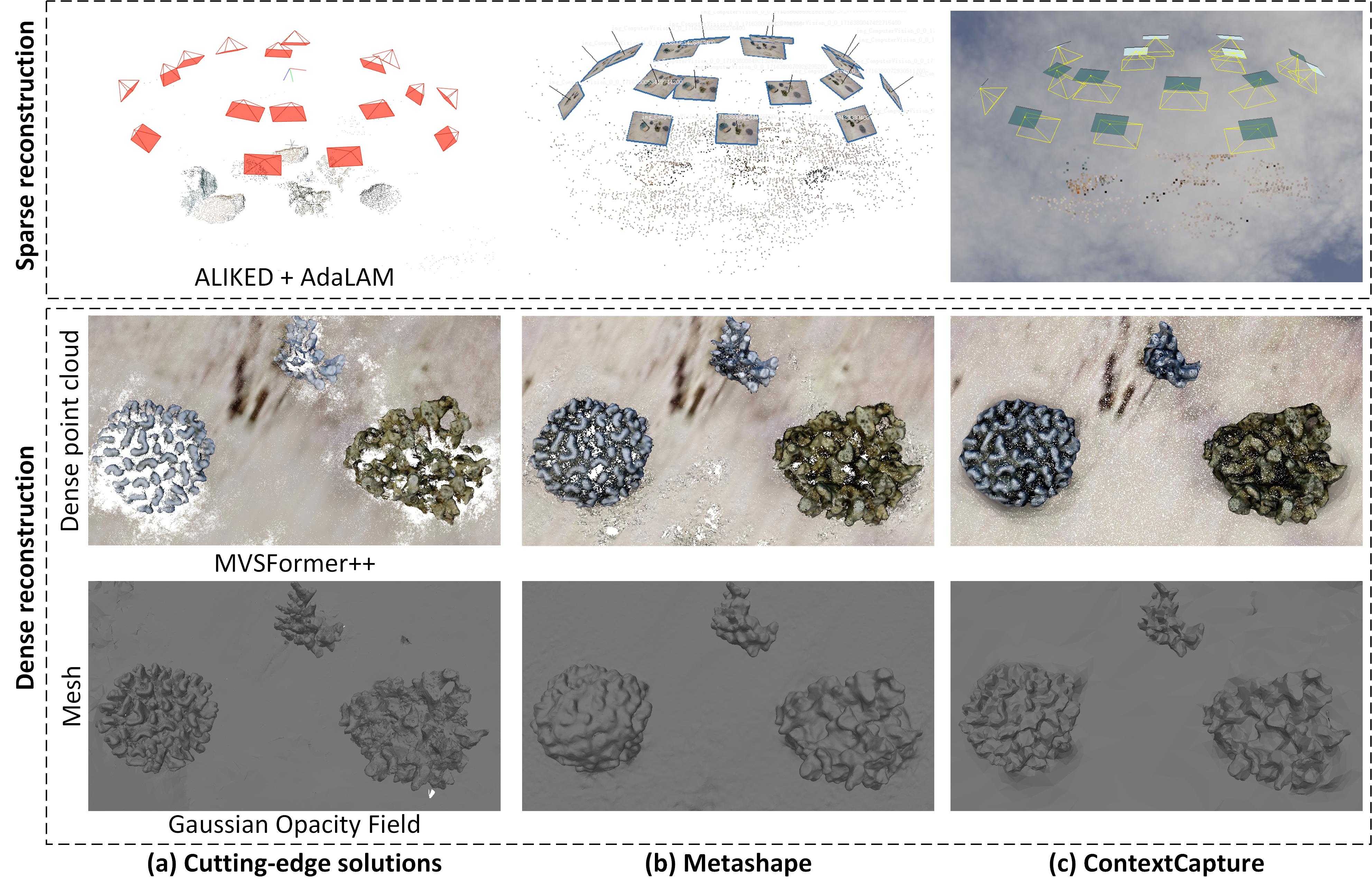}
	\caption{3D reconstruction results of different solutions on the Coral-UE4 (lite) dataset. The SfM reconstruction results shown are screenshots from the software.}
	\label{fig15}
\end{figure*}

To further assess the applicability of recent cutting-edge solutions and provide references for practical engineering applications, we compared two leading commercial 3D reconstruction software packages — Agisoft Metashape and Bentley ContextCapture. These software packages offer functionalities for camera pose estimation and dense reconstruction, and have been used in coral reef reconstruction studies (\cite{bayley2020protocol120,rossi2020detecting18,burns2017comparison122,urbina2021quantifying123}). For camera pose estimation, we compare the performance of the "ALIKED feature with AdaLAM matching" solution (referred to as ALIKED+AdaLAM) as a benchmark. In Metashape, the “\textit{Align Photos}” function is used to achieve this, while in ContextCapture, the “\textit{Aerotriangulation}” process is employed. For dense reconstruction, we employ MVSFormer++ and GOF as reference methods for extracting dense point clouds and mesh models, respectively. In Metashape, the “\textit{Build Dense Cloud}” and “\textit{Build Mesh}” functions are used for creating the dense point cloud and mesh model, whereas in ContextCapture, these tasks are carried out by submitting jobs for the “\textit{3D point cloud}” and “\textit{3D mesh}” reconstruction. The parameters used in Metashape and ContextCapture are presented in Table \ref{tab6}. Figure \ref{fig15} illustrates the reconstruction results obtained from processing the Coral-UE4 (lite) dataset with each of these solutions.

\begin{table*}[ht]
	\caption{Parameter Settings for Metashape and ContextCapture.}
	\centering
	\label{tab6}
	\renewcommand\arraystretch{1.1}
	\begin{tabular}{llll}
		\hline
		Software                        & Function                                    & Parameter                  & Value               \\
		\hline
		\multirow{6}{*}{Metashape}      & \multirow{3}{*}{\textit{Align Photos}}      & \textit{Accuracy}          & \textit{Highest}    \\
		&                                             & \textit{Key point limit}   & \textit{10,000}     \\
		&                                             & \textit{Tie point limit}   & \textit{5,000}      \\
		& \multirow{2}{*}{\textit{Build Dense Cloud}} & \textit{Quality}           & \textit{Ultra high} \\
		&                                             & \textit{Depth filtering}   & \textit{Moderate}   \\
		& \textit{Build Mesh}                         & \textit{Face}              & \textit{High}       \\
		\hline
		\multirow{4}{*}{ContextCapture} & \multirow{2}{*}{\textit{Aerotriangulation}} & \textit{Key point density} & \textit{High}       \\
		&                                             & \textit{Pair selection mode}        & \textit{Exhaustive}          \\
		& \textit{3D point cloud }                             & \textit{Point sampling }            & \textit{1 pixel}             \\
		&\textit {3D mesh}                                     & \textit{Node size}                  & \textit{large}              \\
		\hline
	\end{tabular}
\end{table*}

For SfM reconstruction, as detailed in Section \ref{E2S2}, we evaluate each method based on several metrics such as location accuracy calculated using the simulation dataset. The results are summarized in Table \ref{tab7}. ContextCapture shows the poorest performance, particularly on the Coral-UE4 dataset, where it struggled to align all images and demonstrated significant pose estimation errors. This suggests its difficulties in accurately matching images with weak textures. Additionally, its performance is quite poor with limited data, with fewer than 1000 successfully matched points per image. In contrast, the reconstruction accuracy of Metashape is comparable to that of ALIKED+AdaLAM, but it exhibits significantly lower performance in both \textit{Features} and \textit{Track}. The results depicted in Figure \ref{fig15} reveal that the point cloud generated by ALIKED+AdaLAM predominantly aligns with the coral reef, indicating that the successfully matched features are largely concentrated in the coral reef areas of the images. This concentration contributes to its high \textit{Track} value of this solution. On the other hand, Metashape produces a point cloud that is evenly distributed across the entire scene, while the sparse point cloud of ContextCapture is notably limited, predominantly consisting of points of the ground instead of coral reefs. This difference is likely related to the design of the software. For instance, Metashape may prioritize feature matching across various regions of the image, leading to a more uniform point distribution.

\begin{table}[]
	\caption{Various metrics of the reconstruction results of Coral-UE4 (lite) using different solutions.}
	\centering
	\label{tab7}
	\begin{tabular}{lllll}
		\hline
		&                   & \begin{tabular}[c]{@{}l@{}}ALIKED\\   +AdaLAM\end{tabular} & Metashape & ContextCapture \\
		\hline
		\multirow{6}{*}{\rotatebox[origin=c]{90}{Coral-UE4}}                                                          & \textit{Rate}     & \textbf{66/66}                                                         & \textbf{66/66}     & 63/66          \\
		& \textit{Features} & \textbf{6600.3}                                                        & 1853.6    & 881.3          \\
		& \textit{Points}   & \textbf{41,640}                                                        & 35,742    & 12,944         \\
		& \textit{Track}    & \textbf{10.46}                                                         & 3.42      & 4.29           \\
		& $E_{rep}$   & 1.05                                                          & 0.78      & \textbf{0.57}           \\
		& $E_{loc}$   & 0.76                                                          & \textbf{0.76}      & 7.82           \\
		\hline
		\multirow{6}{*}{\rotatebox[origin=c]{90}{Coral-UE4 (lite)}} & \textit{Rate}     & \textbf{17/17}                                                         & \textbf{17/17}     & 15/17          \\
		& \textit{Features} & \textbf{2150.8}                                                        & 834.2     & 236.3          \\
		& \textit{Points}   & \textbf{10,057}                                                        & 5,064     & 934            \\
		& \textit{Track}    & 3.64                                                          & 2.80      & \textbf{3.79}           \\
		& $E_{rep}$   & 0.69                                                          & 0.76      & \textbf{0.62}           \\
		& $E_{loc}$   & 0.74                                                          & \textbf{0.73 }     & 0.76           \\
		\hline
		\multirow{6}{*}{\rotatebox[origin=c]{90}{Coral-UE5}}                                                          & \textit{Rate}     & \textbf{70/70}                                                         & \textbf{70/70}     & \textbf{70/70}          \\
		& \textit{Features} &\textbf{6785.8 }                                                       & 2723.2    & 2347.7         \\
		& \textit{Points}   & \textbf{54,633 }                                                       & 40,122    & 32,726         \\
		& \textit{Track}    & \textbf{8.69 }                                                         & 4.75      & 5.02           \\
		& $E_{rep}$   & 0.76                                                          & 0.47      & \textbf{0.44 }          \\
		& $E_{loc}$   & 1.18                                                          & 1.18      & \textbf{1.18}           \\
		\hline
		\multirow{6}{*}{\rotatebox[origin=c]{90}{Coral-UE5 (lite)}}   & \textit{Rate}     & \textbf{18/18 }                                                        & \textbf{18/18}     & \textbf{18/18}          \\
		& \textit{Features} & \textbf{2374.4}                                                        & 2062.8    & 907.7          \\
		& \textit{Points}   & 11,574                                                        & \textbf{13,170}    & 3,723          \\
		& \textit{Track}    & 3.69                                                          & 2.82      & \textbf{4.39}           \\
		& $E_{rep}$   & 0.65                                                          & \textbf{0.39}      & 0.49           \\
		& $E_{loc}$   & \textbf{1.19}                                                          & 1.20      & 1.20          \\
		\hline
	\end{tabular}
\end{table}

\begin{figure*}
	\centering
	\includegraphics[width=2.0\columnwidth]{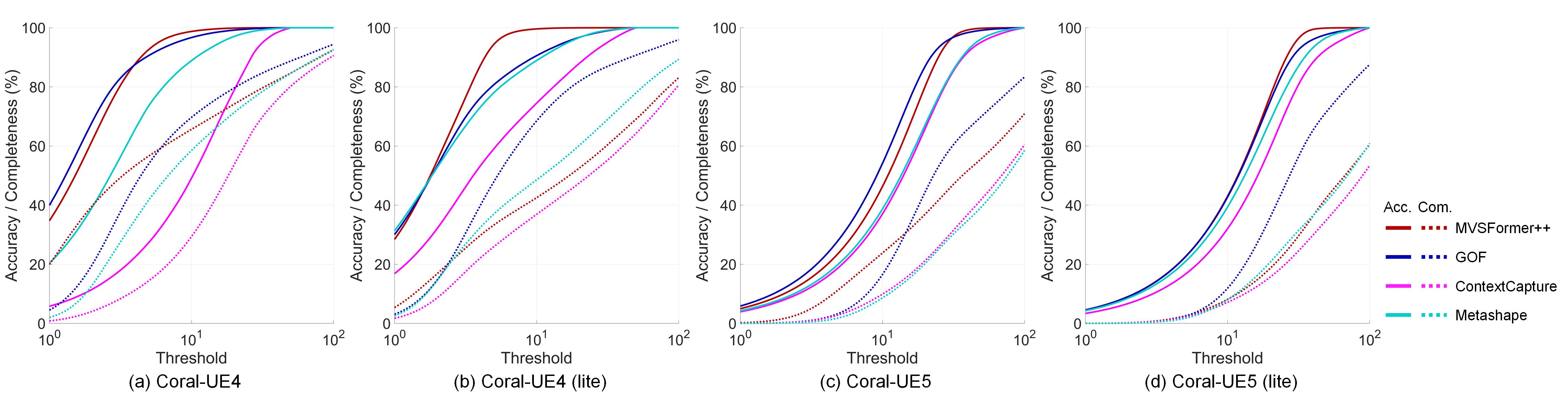}
	\caption{Accuracy and completeness of the reconstruction results for each approach at different distance thresholds.}
	\label{fig16}
\end{figure*}

In terms of dense reconstruction, the results presented in Figure \ref{fig15} indicated that the dense point cloud generated by MVSFormer++ is less complete compared to that produced by commercial software. This incompleteness primarily arises from shadowed areas caused by occlusion of small coral structures. These regions may have been deemed unreliable and thus excluded. Additionally, ContextCapture failed to fully reconstruct the geometric structure of the smaller corals located in the upper and middle portions of the scene. Furthermore, errors are observed along the edges of the corals in ContextCapture's results, likely due to interference from background textures during the dense matching process. In mesh reconstruction, GOF demonstrates exceptional capability in recovering fine details, accurately reconstructing even the small tentacles of the coral reefs. In contrast, Metashape's results are relatively smooth, exhibiting no large, conspicuous errors; however, it sacrifices a degree of finer detail in the process. ContextCapture produces mesh models with lower fidelity, featuring fewer mesh faces and failing to recover the intricate structure of the coral reefs. As detailed in Section \ref{E3S1}, we compute the accuracy and completeness of each method's results, which are shown in Figure \ref{fig16}. Overall, ContextCapture has the lowest accuracy and completeness, particularly in the Coral-UE4 dataset, due to its errors in camera pose estimation. Metashape significantly outperforms ContextCapture in the first two datasets, while its performance in the other two datasets is relatively comparable. In contrast, GOF demonstrates superior accuracy and effectively captures fine details, reaffirming its value for studying coral reef structures.

\subsection{Coral reef metric estimation via dense surface reconstruction}

\begin{figure}
	\centering
	\includegraphics[width=1.0\columnwidth]{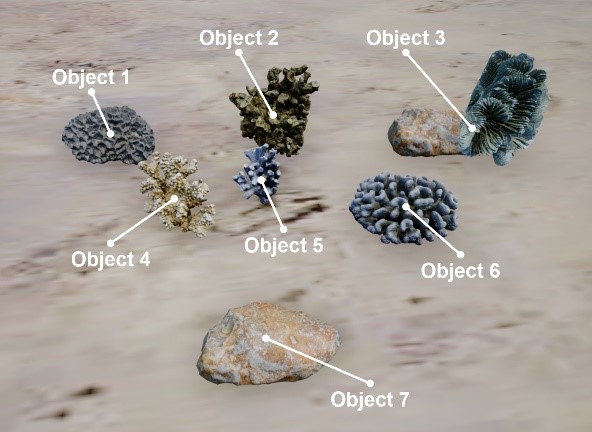}
	\caption{The IDs of objects in the Coral-UE4.}
	\label{fig17}
\end{figure}

\begin{figure*}
	\centering
	\includegraphics[width=2.0\columnwidth]{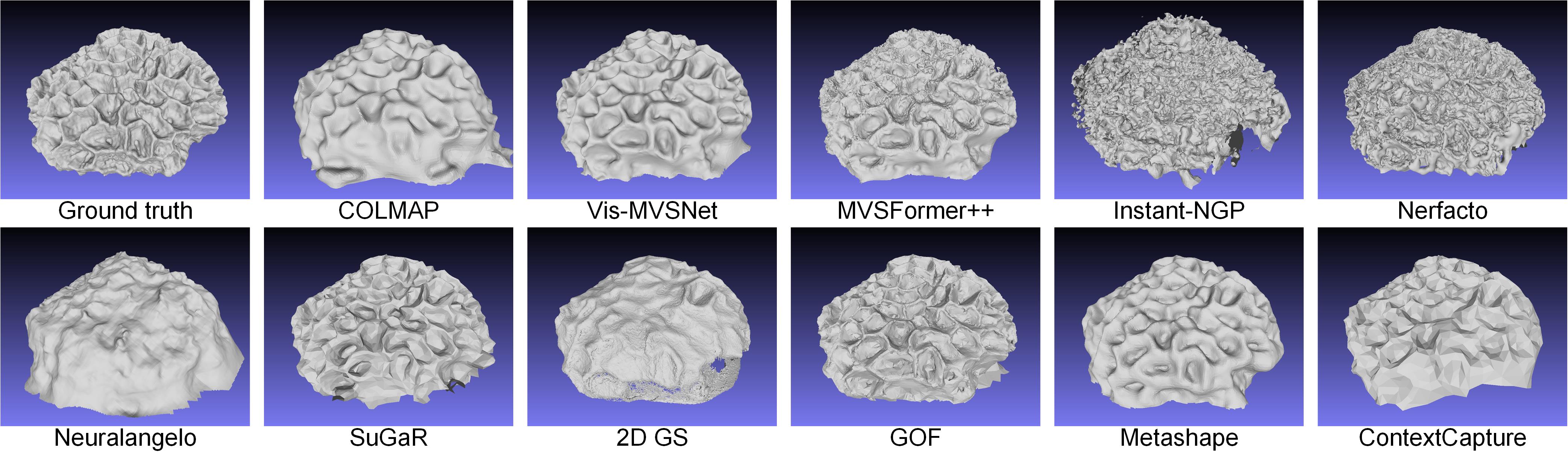}
	\caption{Mesh model of Object 1 in the Coral-UE4 scene generated by different dense reconstruction solutions.}
	\label{fig18}
\end{figure*}

To assess the ecosystem services and functions of corals, surface area and volume are critical 3D metrics (\cite{zawada2019morphological128}). Surface area is significant as it indicates where coral biomass is concentrated and where coral interacts with its environment (\cite{johannes1970method129}). Volume is essential for evaluating the coral reef's capacity to support biodiversity (\cite{urbina2021quantifying123}). Accurate estimation of these metrics is vital for effective coral reef monitoring and conservation. Dense surface reconstruction is a key technology in this process. To assess the effectiveness of different dense reconstruction methods in estimating coral reef metrics, we use the Coral-UE4 dataset as a case study. We produce mesh models of the scene using the approaches involved in Section \ref{E3} and calculate their surface area and volume. As the reconstructed coral reef mesh models are typically not closed, the holes are filled to create a watertight model firstly. The relative error is then calculated as the difference between estimated and true values, divided by the true value, indicating the accuracy of the estimate. Due to significant variation in surface complexity among different objects, the coral reefs in Coral-UE4 are divided into seven separate objects for statistical analysis, numbered 1 to 7, as illustrated in Figure \ref{fig17}. The relative errors for surface area estimation are presented in Table \ref{tab8}, and for volume estimations in Table \ref{tab9}. We also compute the Root Mean Square Relative Error (RMSRE). It provides an overall assessment of the accuracy of each method. Figure \ref{fig18} presents the reconstructed mesh model of Object 1 for qualitative analysis.

\begin{table*}[]
	\caption{Relative error (\%) in surface area estimation of mesh models.}
	\centering
	\label{tab8}
	\begin{tabular}{lllllllll}
		\hline
		Object   ID    & 1     & 2     & 3     & 4     & 5     & 6     & 7     & RMSRE \\
		\hline
		COLMAP         & -34.6 & -34.4 & -58.3 & -16.6 & -26.8 & -45.6 & -10.8 & 35.8  \\
		Vis-MVSNet     & -29.3 & -32.4 & -52.7 & -21.5 & -24.8 & -29.5 & -8.7  & 30.9  \\
		MVSFormer++    & -5.1  & -17.5 & -46.9 & 2.0   & 10.7  & -11.9 & 8.7   & 20.2  \\
		Instant-NGP    & 136.7 & 111.6 & 121.1 & 160.9 & 111.6 & 115.3 & 219.5 & 144.2 \\
		Nerfacto       & 68.8  & 61.4  & 33.3  & 65.5  & 74.9  & 62.5  & 71.7  & 63.9  \\
		Neuralangelo   & -40.8 & -42.0 & -61.7 & -8.4  & -23.7 & -58.0 & -8.6  & 40.2  \\
		SuGaR          & 17.5  & -19.5 & -29.0 & -7.8  & -5.1  & \textbf{-2.9}  & -10.4 & 15.7  \\
		2D GS          & -33.4 & -29.3 & -46.9 & -10.5 & \textbf{-0.3}  & -8.8  & 8.5   & 25.2  \\
		GOF            & \textbf{3.6}   & \textbf{-10.2} & \textbf{-20.7} & \textbf{0.4}   & 3.0   & -6.0  & \textbf{-1.6}  & \textbf{9.2}   \\
		Metashape      & -34.8 & -24.6 & -56.8 & -19.3 & -21.7 & -39.0 & -10.9 & 32.8  \\
		ContextCapture & -29.1 & -29.8 & -52.2 & -7.6  & -17.5 & -36.6 & -5.8  & 29.8  \\
		\hline
	\end{tabular}
\end{table*}

In the estimation of surface area, Instant-NGP and Nerfacto exhibit inaccuracies due to numerous outliers in their point clouds. These outliers result in mesh models with many erroneous surfaces, causing a substantial overestimation of surface area. Even the smoother regions of the surface appeared uneven in their reconstructions. Conversely, other methods struggle to fully capture the intricate details of the coral reefs but generate fewer errors, leading to underestimations of surface areas. Notably, Neuralangelo produces overly smooth mesh models, which leads to a significant underestimation of surface area. Among the evaluated methods, GOF shows the smallest overall error, demonstrating the highest accuracy in surface area estimation. SuGaR and MVSFormer++ also perform well, capturing sufficient detail even on the more complex surfaces of coral reefs, such as Object 1. Among these objects, Object 7 has the simplest surface, being just a rock. The relative error in surface area estimation for this object is generally below 11\% across all methods. Conversely, Object 3 presents the most intricate and challenging surface to reconstruct, with densely folded structures that are difficult to capture accurately. Consequently, all methods show higher errors in estimating its surface area.

\begin{table*}[]
	\caption{Relative error (\%) in volume estimation of mesh models.}
	\centering
	\label{tab9}
	\begin{tabular}{lllllllll}
		\hline
		Object   ID    & 1     & 2     & 3     & 4     & 5     & 6     & 7     & RMSRE \\
		\hline
		COLMAP         & 7.8   & 25.2  & 32.3  & 55.9  & 19.9  & 65.7  & -15.4 & 37.4  \\
		Vis-MVSNet     & \textbf{0.4}   & 10.2  & 21.2  & 17.7  & -5.2  & 22.0  & -9.8  & 14.5  \\
		MVSFormer++    & 1.6   & 15.4  & \textbf{0.4}   & 39.8  & 23.2  & -20.0 & -7.2  & 20.1  \\
		Instant-NGP    & -10.9 & 6.0   & 12.2  & 142.5 & 37.2  & 27.9  & -60.1 & 61.4  \\
		Nerfacto       & -9.6  & -40.4 & -13.7 & -13.2 & -20.7 & -15.3 & -13.4 & 20.5  \\
		Neuralangelo   & 18.1  & 41.3  & 39.4  & 98.1  & 54.9  & 71.3  & 5.5   & 55.2  \\
		SuGaR          & -5.9  & 4.6   & 4.9   & 6.6   & -6.1  & \textbf{1.4}   & \textbf{2.0}   & 4.9   \\
		2D GS          & 3.4   & \textbf{3.5}   & 12.8  & -26.7 & -25.5 & 4.0   & -10.3 & 15.5  \\
		GOF            & -1.3  & 4.9   & -0.6  & \textbf{-2.4}  & -6.4  & 3.2   & -7.0  & \textbf{4.4}   \\
		Metashape      & 5.2   & 22.2  & 21.8  & 48.5  & 41.0  & 56.7  & -6.0  & 34.4  \\
		ContextCapture & -5.1  & -3.9  & 17.9  & 53.0  & \textbf{4.5}   & 49.0  & -8.2  & 28.4  \\
		\hline
	\end{tabular}
\end{table*}

Volume is often overestimated, unlike surface area which tends to be underestimated. This occurs because most reconstructed points are located on the outer ends of the complex structure, making it difficult to capture concave region accurately. This issue is particularly evident in Neuralangelo's results, where the inability to accurately reconstruct the details of the coral's contact with the ground leads to an incorrect merging of the mesh in that region, resulting in a significantly inflated volume estimate. Similarly, Metashape tends to overestimate volume due to excessive expansion of the mesh. On the other hand, cases of underestimated volume usually arise from incomplete reconstructions or erroneous depressions in the mesh. Among these methods, GS-based approaches exhibit the lowest overall error. They differ from NeRF-based methods by offering a more direct and accurate description of local geometry through Gaussian representations. This enables them to perform exceptionally well in scenarios with complex geometric shapes. Deep learning-based MVS methods also perform well overall but face challenges when reconstructing intricate structures, suggesting areas for future improvement. In summary, for accurately estimating the 3D metrics of coral reefs with complex geometric structures, GS-based methods are the most suitable choice.

\section{Discussions}\label{}

\subsection{Accuracy and robustness of camera pose estimation}

The primary objective in camera pose estimation is to compute accurate and robust camera parameters for all images within a dataset. Our experiments (Section \ref{E2}) show that not all cutting edge matching methods guarantee improved performance across all metrics, but several recent techniques demonstrate clear benefits in specific scenarios. Overall, certain emerging techniques offer compelling advantages.

For example, when using SIFT as the feature detector, LightGlue achieves more accurate overall reconstructions, notably on the Coral UE5 (lite) dataset. Although LightGlue sometimes produces a slightly higher reprojection error compared to the traditional ratio test, this does not directly correlate with object space accuracy. LightGlue's contribution led to a more complete and globally consistent structure, making it a superior choice for the overall reconstruction pipeline. Similarly, ALIKED proves to be an effective feature detector, offering a level of scale invariance that many deep learning based descriptors struggle to match. Its robustness greatly increases the success rate of image alignment, positioning it as a strong alternative.

To improve the success rate and quality of the subsequent dense reconstruction phase in the overall workflow, it is important to align as many images as possible during the SfM stage. Our findings across different datasets reveal that so long as the images are successfully aligned, the final accuracy tends to converge to a comparable level. For instance, multiple pipelines achieve millimeter-level error of the check point on Coral-2018 and similar pose errors on Coral-UE5. These suggest that the current bottleneck lies not in the precision of feature localization but in the robustness of feature matching. Specifically, the ability to establish reliable correspondences, especially under challenging conditions. The comparison with commercial software in Section \ref{software} supports this view, where ContextCapture's failure to align all images on the Coral-UE4 dataset led to significantly larger errors.

The promising performance of methods like SuperGlue and LightGlue underscores the potential of deep learning-based matching techniques. These methods leverage broader image context to generate more reliable correspondences than traditional approaches. This success suggests that further exploration of these avenues is valuable. Furthermore, exploring the integration of prior knowledge, like known image sequences or approximate geolocation, offers another potential pathway to bolster alignment robustness, especially within large or complex scenes.

\subsection{ Dense surface reconstruction with enhanced quality and efficiency}

Regarding dense surface reconstruction, our experiments highlight the transformative impact of recent advances like NeRF and 3D GS. These techniques have spurred the development of novel methods capable of achieving high-quality reconstructions, capturing intricate geometric details and realistic appearances. Concurrently, deep learning-based MVS methods, which integrate deep learning components into typical MVS pipelines, deliver strong performance with favorable runtimes. As technology evolves, we anticipate further breakthroughs in both approaches.

In practical applications, particularly for large-scale surveys common in environmental monitoring (e.g., coral reef mapping), the trade-off between reconstruction quality and computational efficiency is a critical consideration. Based on the results presented in Section \ref{E3}, deep learning-based MVS methods, particularly MVSFormer++, achieve an excellent balance. They offer a compelling general-purpose solution, delivering high-quality results with reasonable computational costs. While methods like GOF can achieve outstanding accuracy and completeness, their significant computational cost may render them less practical for extensive areas.

However, for coral reef 3D reconstruction specifically, a single method may not be optimal for the entire scene. Reef environments often contain diverse structures: complex coral colonies requiring high fidelity, alongside simpler areas like rock formations or sandy patches where extreme detail is less critical. This heterogeneity suggests a potential hybrid strategy. For example, one could employ an efficient method like MVSFormer++ for broad-scale reconstruction of the entire survey area. Subsequently, specific regions of interest containing intricate or scientifically significant coral structures could be selectively reconstructed using a high-fidelity but more computationally intensive method like GOF. The resulting high-detail local models could then be integrated into global reconstruction. This offers a practical pathway to balance overall quality and efficiency. Developing robust methodologies for planning such hybrid reconstructions, segmenting regions for differential processing, and seamlessly fusing the results from different methods represents a valuable direction for future research, potentially leading to optimized workflows tailored for complex, large-scale 3D modeling tasks.

\subsection{Issues and future studies}

This paper provides a review of the current cutting-edge solutions for camera pose estimation and dense reconstruction, evaluates these methods through experiments on both real-world and synthetic data, and compares them with commonly used commercial software, offering valuable insights and recommendations. Although significant progress has been made in 3D reconstruction, several challenges remain unresolved, particularly in the context of underwater coral reef scenes. The improvement of underwater 3D reconstruction technology can be pursued through the following directions:

(1) Optimization and trade-offs in performance and computational costs. 3D reconstruction, essential for applications such as measurement, mapping, and monitoring, inherently prioritizes accuracy. Many studies focus on improving algorithmic performance, often at the expense of computational costs. However, as image resolution increases and survey areas expand, processing costs also grow significantly and cannot be overlooked. The ideal objective is to implement a 3D reconstruction solution that can accurately capture scene geometry and texture while utilizing resources efficiently, including runtime, computational load, and equipment costs. For example, MVSFormer++ exemplifies a commendable balance by achieving high accuracy with minimal processing time. In contrast, NeRF-based and GS-based methods, despite their exceptional performance (e.g., GOF), often struggle with significant computational demands, such as high memory usage, extended training times, and limited generalization across different scenes. These challenges constrain their practical application. Besides algorithmic advancements, balancing performance and cost can be achieved through the integration of multiple approaches. For instance, in dense reconstruction of extensive coral reef areas, MVSFormer++ can quickly produce high-quality dense models of the entire area, while methods like GOF can be applied to specific regions of interest to enhance detail and reliability; adopting strategies such as a coarse-to-fine approach can further optimize the balance between performance and cost. Moreover, recent fully feed forward, multi-task geometry predictors such as VGGT (\cite{wang2025vggt149}), $\pi^{3}$ (\cite{wang2025pi150}) and Dense3R (\cite{fang2025dens3r151}) accept unposed or unordered image sets and directly infer multiple geometric outputs in a single forward pass without requiring traditional optimization or explicit geometric priors. This unified end-to-end paradigm offers promising robustness, scalability, and efficiency for future underwater 3D reconstruction workflows.

(2) Evaluation metrics for coral reef 3D reconstruction. Currently, algorithms for evaluating reconstruction quality are underdeveloped, and progress in this area is relatively slow. Coral reefs present a unique and challenging scenario due to their intricate, densely packed structures and varying topography. Additionally, as coral reefs grow slowly—only a few centimeters per year—the reconstruction accuracy needs to achieve millimeter-level precision (\cite{zhong2023combining53}). Therefore, accurate assessment of reconstruction results is crucial and necessitates the development of more representative evaluation metrics for quantifying shape reconstruction analysis results (\cite{chen2019learning99,figueira2015accuracy134}). Future research on evaluation methods should consider a comprehensive range of factors, including global and local perspectives, geometric and radiometric aspects, as well as accuracy and uncertainty. In ecological monitoring, it is crucial to focus on metrics that reflect coral reefs’ ecological functionality, such as biomass, diameter, height, and surface roughness. These indicators can be used to evaluate how well 3D reconstructions capture ecological features, supporting the conservation and management of coral reef ecosystems.

(3) Point Cloud Processing. In 3D reconstruction, dense methods often generate point clouds as intermediate outputs. Efficient processing and utilization of these point clouds represent a challenge. Currently, point clouds are commonly used in reconstruction either as references or initializations for subsequent processes, or more commonly, for reconstructing scene surfaces for further applications. However, this problem is technically ill-posed, as there are infinitely many continuous surface solutions for a given set of discrete points (\cite{huang2024surface100}). The problem is further complicated by the presence of potential errors, uneven point distribution, missing points, and even incorrect points within the reconstructed data. These challenges are particularly acute in complex scenarios such as coral reefs, where intricate structures and occlusions make reliable surface reconstruction exceptionally difficult. As demonstrated in Figure \ref{fig9}, the dense point clouds and mesh models produced by MVS methods reveal that regions with missing points or anomalies in the point clouds lead to significant distortions in the reconstructed meshes. Such anomalies not only compromise the positional accuracy of the mesh but also negatively impact surface reconstruction by affecting the estimation of surface normals, as the use of surface normals is crucial for successful surface reconstruction (\cite{huang2024surface100}). For instance, the inaccurate normal estimation in point clouds generated by Instant-NGP leads to a mesh that is not smooth but rather rugged. Consequently, future research should focus on not only improving point cloud quality and optimizing mesh reconstruction strategies but also on enhancing normal estimation approaches. Recent advancements have shown the application of deep learning methods achieving some success in this area, and future efforts may consider leveraging multi-view image information for further improvements.

(4) Addressing the effects of water on imaging. As light propagates through water, it is subject to both absorption and scattering, which significantly degrade image quality. The absorption of light varies across wavelengths, leading to color distortion, while scattering introduces blur and haze (\cite{akkaynak2019sea141}). In addition, underwater images may be degraded by turbidity and water turbulence, further complicating the imaging process. To mitigate these issues, current approaches typically involve color calibration or advanced image enhancement techniques, including generative adversarial networks (GANs) (\cite{yu2019underwater126}) and diffusion models (\cite{tang2023underwater127}). However, camera color calibration typically targets individual images, making it time-consuming and less effective for multi-view datasets. While image enhancement techniques can produce visually convincing results, they often lack physical accuracy, resulting in inconsistencies between color information and geometric structures. The NeRF framework, however, offers a promising solution to the challenges of light scattering in underwater environments. NeRF methods utilize volumetric rendering, allowing them to model both the geometry and the medium of a scene. For instance, Levy et al. introduced SeaThru-NeRF (\cite{levy2023seathru124}), which integrates a scattering image formation model into the NeRF rendering equations to separate backscatter components from the scene. Similarly, Li et al. developed WaterSplatting (\cite{li2024watersplatting125}), employing 3D Gaussian Splatting to explicitly represent the scene's geometry while utilizing a separate volumetric field to capture the water. These methods facilitate the interpretation and modeling of light propagation, scattering, and absorption in underwater environments from a 3D perspective, while optimizing color and geometry. Consequently, they improve consistency between color and geometric information across varying underwater conditions, supporting high-quality 3D reconstruction. Looking ahead, combining these techniques with surface reconstruction methods like SuGaR and GOF could lead to significant advancements in underwater 3D reconstruction.

(5) Multi-source data fusion for underwater 3D reconstruction. In addition to photogrammetry, other sensing approaches such as sonar and LiDAR have been utilized for underwater 3D reconstruction. Sonar enables rapid, large-scale surveys and is useful for broad-area mapping, though its resolution is insufficient for capturing fine-scale structures when compared to image-based methods, which can achieve millimeter-level accuracy. Airborne LiDAR has been widely used in coastal and nearshore ocean mapping and can recover both the ocean surface and seabed topography (\cite{massot2015optical144}). Nevertheless, it is limited by relatively low resolution and is susceptible to water surface refraction. In contrast, underwater LiDAR can operate in close proximity to the scene, reducing water-related distortion and allowing for direct distance measurement (\cite{yang2022underwater142,mcleod2013autonomous143}). This makes underwater LiDAR particularly valuable as a complement to image-based approaches, as its depth information aids in modeling light propagation through water. However, underwater LiDAR remains relatively underdeveloped and is mainly used as a supplementary tool. With further advancements, particularly toward performance levels comparable to terrestrial LiDAR, it holds great promise for improving the reconstruction of complex structures such as coral reefs, offering enhanced accuracy and robustness through data fusion.

(6) Demand for datasets and the application of simulators. To advance the study and evaluation of algorithms, it is crucial to have both appropriate metrics and suitable datasets. However, due to the limited distribution of coral reefs and the challenges of underwater photography, datasets for coral reef 3D reconstruction are rare. Even available datasets often lack ground-truth 3D information of the scenes, making it difficult to rigorously assess reconstruction accuracy. This highlights the urgent need for relevant datasets. While obtaining ground-truth data in real-world coral reefs is challenging, simulation offers a viable solution. Recent advance in simulators, particularly in autonomous driving, illustrate this potential (\cite{rong2020lgsvl101,hu2023simulation102}). For example, simulators like CARLA enable extensive testing by generating synthetic scenarios, which can also benefit 3D reconstruction (\cite{dosovitskiy2017carla103}). Synthetic data has proven effective for training models in SuperPoint and for evaluating algorithms in NeRF and Gaussian Splatting. Therefore, it is both reasonable and feasible to develop simulation environments and synthetic datasets for coral reef 3D reconstruction, as demonstrated by the datasets generated using AirSim in this study. Several works have contributed to this area. GEODT (\cite{nakath2022optical146}), an optical digital twin for underwater environments, supports the development and validation of refractive photogrammetric algorithms. POSER (\cite{menna2024poser147}) was proposed for teaching and training underwater photogrammetry, allowing users to explore various acquisition strategies. Potokar et al. proposed HoloOcean (\cite{potokar2022holoocean136}), an underwater robotics simulator based on Unreal Engine 4 that supports the generation of photorealistic underwater imagery. Amer et al. introduced Unav-Sim (\cite{amer2023unav137}), which enables the simulation of different underwater scenarios and models. In March 2025, Song et al. presented OceanSim (\cite{song2025oceansim135}), a high-fidelity simulator designed to improve physics-based underwater sensor modeling. Future development should consider various factors, including coral morphologies, water quality, and lighting conditions, ensuring the simulator achieves high fidelity to closely mimic real-world environments. Despite advancements, inherent discrepancies remain between simulated and real-world environments, particularly concerning factors such as lighting and texture. Consequently, research in simulation-to-real transfer and reality gap modeling is essential to effectively address these shortcomings (\cite{daza2023sim104}).

\section{Conclusions}\label{}

In this study, we conduct a systematic review and experimental evaluation of the current cutting-edge 3D reconstruction solutions for coral habitat modeling using underwater images, with a focus on camera pose estimation and dense reconstruction techniques. We elaborate on how the latest advancements in photogrammetric computer vision and deep learning technologies can be applied to high-resolution underwater 3D reconstruction, providing finer comprehensive guidance for seabed reef mapping practices.

For camera pose estimation, we focus on techniques related to feature extraction, feature matching, and SfM reconstruction. Our evaluation highlighted that deep learning-based local feature extraction and feature matching methods have significantly outperformed traditional hand-crafted approaches, especially under conditions of underwater weak texture or variable illumination. Nonetheless, challenges remain in achieving rotational invariance. Learning-based SfM frameworks have demonstrated promising results, though they do not exhibit a clear advantage over conventional incremental SfM frameworks. As for dense reconstruction, we explore four categories of solutions: traditional MVS, deep learning-based MVS, NeRF-based methods, and GS-based methods. Deep learning-based MVS methods showed the best overall performance, excelling in both accuracy and efficiency, thus emerging as the most practical coral reef modeling choice. NeRF-based and GS-based methods displayed varied results, with several methods proving unsuitable for coral reef scenes. However, among these, GOF achieved the highest accuracy and most favorable outcomes, indicating strong potential for future development. Furthermore, a comparison with commercial software revealed that cutting-edge solutions are not only competitive but may also surpass existing options. Building on these findings, we discuss the existing challenges and outline potential research directions about coral seabed reconstruction in five key areas: performance versus cost, evaluation metrics, point cloud processing, light scattering mitigation, and dataset development. Our future work will focus on advancing this domain and refining research resources, including datasets.

Overall, this study aims to inform coral preservation and monitoring researchers and practitioners about the available solutions for underwater reef 3D reconstruction with images, thereby supporting ongoing efforts in coral reef system subsea remote sensing monitoring and conservation and enabling more detailed research and assessment of coral reef ecosystems' roles in the context of future global climate warming. Thus, this work holds considerable urgency.

%
%




\section*{Acknowledgements}

The data for this study was obtained from the Moorea IDEA project. This work was supported by the National Science Fund for Distinguished Young Scholars (grant number 62425102); the U.S. National Science Foundation (grant number OCE 2224354 and earlier awards) for the Moorea Coral Reef LTER; the Gordon and Betty Moore Foundation in support of the MCR LTER; the Italian Minister of University and Research (grant number PNRA18 00263- B2); the National Natural Science Foundation of China (NSFC- 41901407); the Institute of Theoretical Physics, ETH Zurich. We especially would like to thank Prof. Matthias Troyer for his financial and scientific support provided. Special thanks are extended to the ETH Zurich Photogrammetry and Remote Sensing Group. The numerical calculations in this paper have been done on the supercomputing system in the Supercomputing Center of Wuhan University.


\bibliographystyle{cas-model2-names}

\bibliography{cas-refs}



\clearpage
\onecolumn
\setcounter{table}{0}
\renewcommand{\thetable}{A.\arabic{table}}
\section*{Appendix A}
Details of the involved methods.

\begin{table*}[pos=h]
	\caption{The list of free and open-source codes of feature extraction methods involved in the comparative experiments of this paper.}
	\centering
	\label{tab1a}
	\renewcommand\arraystretch{1.1}
	\begin{adjustbox}{scale={0.81}, center}
	\begin{tabular}{lll}
		\hline
		Name       & Category                                       & Codes                                                                                  \\
		\hline
		SIFT  (\cite{lowe2004distinctive13})     & A   hand-crafted local feature detector        & \begin{tabular}[c]{@{}l@{}}Refer   to the relevant code in the OpenCV library at\\  https://github.com/opencv/opencv\end{tabular} \\
		KAZE (\cite{alcantarilla2012kaze1})      & A   hand-crafted local feature detector        & \begin{tabular}[c]{@{}l@{}}Refer   to the relevant code in the OpenCV library at\\  https://github.com/opencv/opencv\end{tabular} \\
		SuperPoint (\cite{detone2018superpoint8}) & A   deep learning-based local feature detector & https://github.com/magicleap/SuperPointPretrainedNetwork                               \\
		R2D2 (\cite{revaud2019r2d2_17})      & A   deep learning-based local feature detector & https://github.com/naver/r2d2                                                          \\
		DISK  (\cite{tyszkiewicz2020disk24})     & A   deep learning-based local feature detector & https://github.com/cvlab-epfl/disk                                                     \\
		ALIKED  (\cite{zhao2023aliked28})   & A   deep learning-based local feature detector & https://github.com/Shiaoming/ALIKED                                                    \\
		DeDoDe  (\cite{edstedt2024dedode50})   & A   deep learning-based local feature detector & https://github.com/Parskatt/DeDoDe                                                    \\
		\hline
	\end{tabular}
\end{adjustbox}
\end{table*}

\begin{table*}[pos=h]
	\caption{The list of free and open-source codes of feature matching methods involved in the comparative experiments of this paper.}
	\centering
	\label{tab2a}
	\renewcommand\arraystretch{1.1}
	\begin{adjustbox}{scale={0.8}, center}
	\begin{tabular}{lll}
		\hline
		Name      & Category                                        & Codes                                                   \\
		\hline
		AdaLAM (\cite{cavalli2020adalam5})   & A   hand-crafted feature matching method        & https://github.com/cavalli1234/AdaLAM                   \\
		SuperGlue (\cite{sarlin2020superglue19}) & A   deep learning-based feature matching method & https://github.com/magicleap/SuperGluePretrainedNetwork \\
		LightGlue (\cite{lindenberger2023lightglue41}) & A   deep learning-based feature matching method & https://github.com/cvg/LightGlue                        \\
		LoFTR (\cite{sun2021loftr21})    & A   detector-free local feature matching method & https://github.com/zju3dv/LoFTR                        \\
		\hline
	\end{tabular}
\end{adjustbox}
\end{table*}

\begin{table*}[pos=h]
	\caption{The list of free and open-source codes of dense reconstruction methods involved in the comparative experiments of this paper.}
	\centering
	\label{tab3a}
	\renewcommand\arraystretch{1.1}
	\begin{adjustbox}{scale={0.8}, center}
	\begin{tabular}{lll}
		\hline
		Name         & Category                           & Codes                                                       \\
		\hline
		COLMAP (\cite{schonberger2016structure20})      & A   traditional MVS method         & https://github.com/colmap/colmap                            \\
		Vis-MVSNet (\cite{zhang2023vis27})  & A   deep learning-based MVS method & https://github.com/jzhangbs/Vis-MVSNet                      \\
		MVSFormer++ (\cite{cao2024mvsformer++52}) & A   deep learning-based MVS method & https://github.com/maybeLx/MVSFormerPlusPlus                \\
		Instant-NGP (\cite{muller2022instant15}) & A   NeRF-based method              & https://github.com/NVlabs/instant-ngp                       \\
		Nerfacto (\cite{tancik2023nerfstudio22})    & A   NeRF-based method              & https://github.com/nerfstudio-project/nerfstudio            \\
		Neuralangelo (\cite{li2023neuralangelo12}) & A   NeRF-based method              & https://github.com/NVlabs/neuralangelo                      \\
		SuGaR (\cite{guedon2024sugar30})        & A   GS-based method                & https://github.com/Anttwo/SuGaR                             \\
		2D GS (\cite{huang20242d32})       & A   GS-based method                & https://github.com/hbb1/2d-gaussian-splatting               \\
		GOF (\cite{yu2024gaussian86})          & A   GS-based method                & https://github.com/autonomousvision/gaussian-opacity-fields\\
		\hline
	\end{tabular}
\end{adjustbox}
\end{table*}

\begin{table}[pos=h]
	\caption{The list of software involved in the comparative experiments of this paper.}
	\centering
	\label{tab4a}
	\renewcommand\arraystretch{1.1}
	\begin{adjustbox}{scale={0.8}}
	\begin{tabular}{lll}
		\hline
		Name           & Description   & Links                                      \\
		\hline
		\begin{tabular}[c]{@{}l@{}}COLMAP \\(\cite{schonberger2016structure20})\end{tabular}         & \begin{tabular}{p{7.5cm}}A   software that performs SfM and MVS for 3D reconstruction from images.  \end{tabular}    & https://github.com/colmap/colmap           \\
		\begin{tabular}[c]{@{}l@{}}DF-SfM \\(\cite{he2024detector66})\end{tabular}        & \begin{tabular}{p{7.5cm}}A   detector-free SfM framework using detector-free matchers to avoid the early   determination of keypoints, while solving the multi-view inconsistency issue.\end{tabular}   & https://github.com/zju3dv/DetectorFreeSfM  \\
		\begin{tabular}[c]{@{}l@{}}VGG-SfM \\(\cite{wang2023visual90})\end{tabular}       & \begin{tabular}{p{7.5cm}}A   SfM program that leverages deep learning techniques to improve the accuracy   and robustness of 3D reconstruction from multiple images. \end{tabular}  & https://github.com/facebookresearch/vggsfm \\
		Metashape      & \begin{tabular}{p{7.5cm}}A   commercial photogrammetry software that provides advanced 3D modeling and   reconstruction capabilities from photographs. \end{tabular}     & https://www.agisoft.com/                   \\
		ContextCapture & \begin{tabular}{p{7.5cm}}A   commercial photogrammetry software that generates highly accurate 3D models   and geospatial data from photographs and laser scans.  \end{tabular} & https://bdn.bentley.com/product/2474      \\
		\hline
	\end{tabular}
\end{adjustbox}
\end{table}

\end{document}